\documentclass{article} % For LaTeX2e
\usepackage{iclr2027_conference,times}

\usepackage{amsmath,amsfonts,bm}

\def\eqref#1{equation~\ref{#1}}
\def\1{\bm{1}}

\DeclareMathAlphabet{\mathsfit}{\encodingdefault}{\sfdefault}{m}{sl}
\SetMathAlphabet{\mathsfit}{bold}{\encodingdefault}{\sfdefault}{bx}{n}

\usepackage{multirow}
\usepackage{booktabs}
\usepackage{makecell}
\usepackage[table]{xcolor}
\usepackage{graphicx}
\usepackage{hyperref}
\usepackage{url}
\usepackage{graphicx}
\usepackage{booktabs}
\usepackage{tabularx}
\usepackage{makecell, adjustbox}
\usepackage{booktabs}
\usepackage{graphicx}
\usepackage{pifont}

\usepackage{graphicx}
\usepackage{subcaption}
\usepackage[most]{tcolorbox}
\usepackage{listings}
\usepackage{xcolor}
\usepackage{pifont}

\usepackage{listings}
\usepackage[most]{tcolorbox}

\title{WorldSolver: Can LLM Agents Simulate the Physical Dynamics via Solver Generation?}

\author{
{\bfseries
Siru Jiang$^{1}$ \quad
Yongzhe Lyu$^{2}$ \quad
Shuo Lu$^{1}$ \quad
Yubin Wang$^{3,\dagger}$ \quad
Yuxiang Zhang$^{3}$
}
\\
{\bfseries
Yue Liao$^{3}$ \quad
Bin Wang$^{3}$ \quad
Jian Liang$^{1,\ddagger}$ \quad
Tieniu Tan$^{1}$
}
\\[0.6em]
\normalfont
$^{1}$NLPR \& MAIS, CASIA
\qquad
$^{2}$PKU
\qquad
$^{3}$Huawei Noah's Ark Lab
\\[0.35em]
\small
$^\ddagger$Corresponding author.
\qquad
$^\dagger$Project lead.
}

\iclrfinalcopy % Uncomment for camera-ready version, but NOT for submission.
\begin{document}

\maketitle
\fancyhead{}
\renewcommand{\headrulewidth}{0pt}

\begin{abstract}
LLM-based agents are increasingly advancing scientific and engineering problem solving, with physics simulation emerging as a challenging yet practical testbed for reproducing complex physical phenomena with application in embodied AI, games and films.
As the workhorse of such simulation, a solver computes how the state of a dynamic system evolves over time.
Building such solvers requires physical understanding to identify appropriate models, mathematical reasoning to formulate the underlying dynamics, and software engineering to implement them as executable code, yet this capability of LLM agents remains underexplored.
To this end, we introduce \textbf{WorldSolver}, a benchmark of 168 simulation tasks derived from physical phenomena in 61 classic computer graphics papers, spanning 7 physical domains.
Each task contains a code scaffold that provides a fixed simulation environment for the scene, with the solver implementation left for the agent to complete.
Specifically, we evaluate them along three dimensions: \textbf{Execution Checks} for successful execution, \textbf{Visual Fidelity} for reproducing the intended dynamic behavior in the rendered simulation, and \textbf{Physical Plausibility} for physics-grounded verification of the generated dynamics.
Experiments on frontier agents reveal that producing executable solvers is difficult itself, and satisfying visual and physical correctness is even harder.
GPT-5.6-Sol and Claude-Opus-5 perform comparatively better than the other evaluated agents, yet achieve overall scores of only 48.7\% and 46.7\%, respectively.
WorldSolver is an early step toward agentic solver generation, and we hope it helps drive progress toward agents that can faithfully simulate the dynamic physical world.
Code is available at \url{https://github.com/sirujiang/WorldSolver}.
\end{abstract}

\section{Introduction}
Large language model (LLM) agents are increasingly capable of tackling complex real-world tasks~\citep{huang2024understanding,guo2024large}, spanning from software engineering~\citep{jimenez2024swebench,yang2024sweagent,chi2026gamedevbench,luo2026gamecraftbench} to scientific discovery~\citep{huang2024mlagentbench,majumder2024discoverybench,somasekharan2025cfdllmbench}.
Recent studies have shown growing interest in physically realistic simulation of 3D worlds, with broad applications in embodied AI, games, and films~\citep{liu2026simworlds,wang2025embodiedgen}.
A line of work focuses on static scene construction through asset composition and spatial layout~\citep{hu2024scenecraft,xia2025scenegenagent}.  
Another explores dynamic scene generation to reproduce physical phenomena and their temporal evolution~\citep{zhang2026gsagent,lv2026physagent,xie2026physcodebench,wang2026simuscene}.

As the workhorse of such simulation, \textit{solver translates the governing equations of a physical model into a numerical procedure that evolves the simulated system over time}~\citep{baraff2023large,stam1999stable}.
Building them requires physical understanding
to identify appropriate models, mathematical reasoning to formulate the underlying dynamics, and software engineering to implement them as executable code~\citep{guendelman2003nonconvex}.
For example, cloth solver handles stiff elastic dynamics and frictional contact at high degrees of freedom~\citep{li2018animplicit}, while fluid solver instead focuses on incompressibility enforcement and free-surface tracking~\citep{enright2002animation}.

\vspace{-3mm}
However, existing works target dynamics scene generation either produce animations in a solver-free manner~\citep{wang2026simuscene} or rely on existing simulation engine~\citep{zhang2026gsagent,xie2026physcodebench,liu2026simworlds}.
In contrast, scientific-computing benchmarks evaluate numerical solver generation for prescribed governing equations, where the physical model to be solved is given in advance~\citep{li2026codepde,hang2026pdeagentbench,somasekharan2025cfdllmbench}.
Neither setting evaluates the full capability required for physical dynamics simulation, spanning physical understanding for model selection, mathematical reasoning for numerical formulation, and software engineering for executable implementation.
Together, they motivate our question:
\par\smallskip
{\centering
\bfseries\itshape
Can LLM agents turn described physical phenomena into dynamics simulation solvers?
\par}
\smallskip
\noindent

\begin{figure}
    \centering
    \includegraphics[width=1\linewidth]{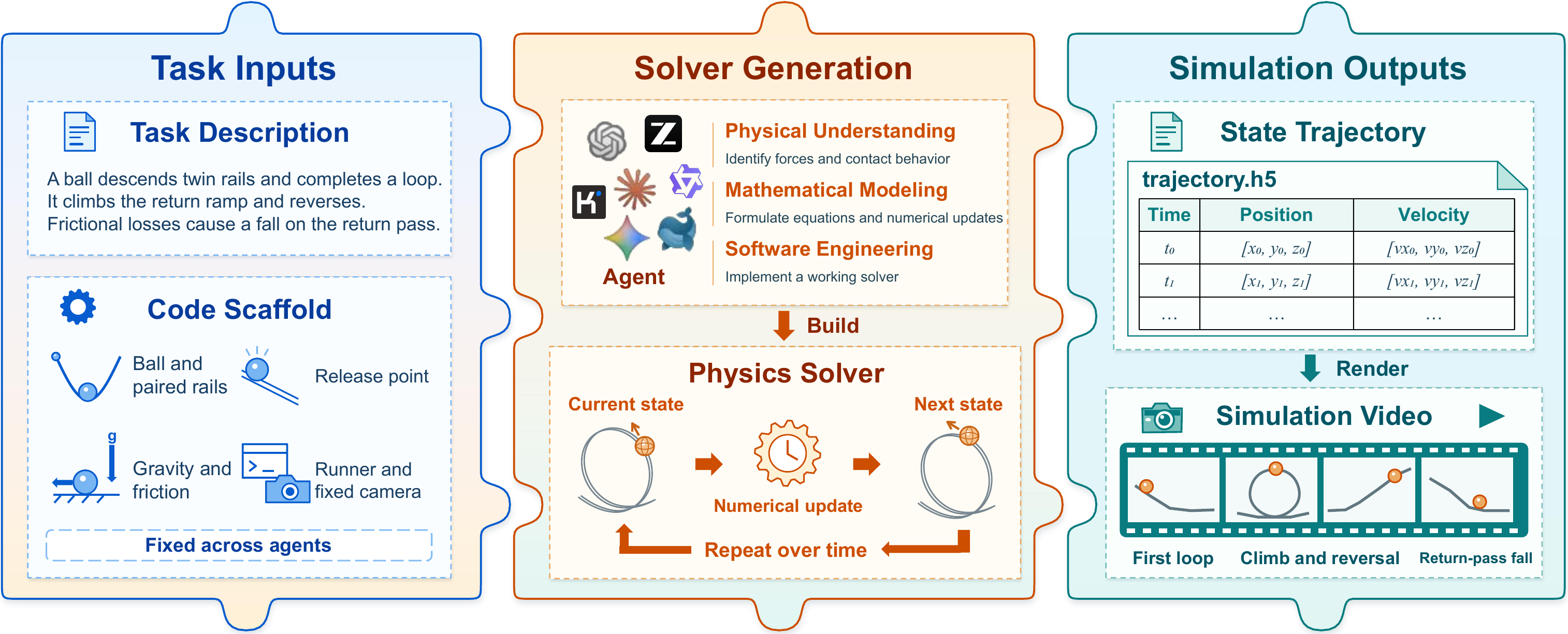}
    \caption{Example of a WorldSolver task.
    Given a task description and a fixed code scaffold, an agent constructs a physics solver leveraging physical understanding, mathematical modeling, and software engineering. 
    The solver advances the system state over time, producing a state trajectory that is rendered into a simulation video. 
    Scene geometry, physical inputs, and rendering settings remain fixed across agents.}
    \label{fig:3stages}
\end{figure}

In this work, we present \textbf{WorldSolver}, a benchmark for evaluating LLM agents' ability to generate solvers for physical dynamics simulation tasks.
As shown in Figure~\ref{fig:3stages}, WorldSolver isolates solver generation with a fixed code scaffold and evaluates the generated physical dynamics through the state trajectory and rendered video outputs.
To ensure a challenging yet physically realistic setting, we curate physical phenomena from classic computer graphics literature and construct simulation tasks around them~\citep{guendelman2003nonconvex,kaufman2005fast,li2018animplicit}.
Given a description of the target phenomenon, the agent implements a solver within a fixed code scaffold that specifies the simulation environment, including scene geometry, physical parameters, external inputs, and rendering configuration.
Once executed, \textit{the solver produces not only a rendered video but also a state trajectory that exposes the underlying physical dynamics through evolving quantities over time}, such as position, velocity, and forces.

Therefore, we evaluate each solver through three protocols: Execution Checks for successful and complete execution, Visual Fidelity for reproducing the intended dynamic behavior in the rendered video, and Physical Plausibility for physically-grounded verification of the generated trajectory.
Table~\ref{tab:benchmark_comparison} summarizes how WorldSolver differs from existing benchmarks in task formulation and evaluation.
Evaluation on the WorldSolver benchmark shows that generating executable solvers from described physical phenomena remains challenging even for frontier coding agents. 
GPT-5.6-Sol and Claude-Opus-5 perform best, yet reach overall scores of only 48.7\% and 46.7\%, while most other agents hover around 20\%.
Our contributions are threefold:

\begin{itemize}
    \item  To the best of our knowledge, we are the first to formulate physical dynamics solver generation as an agentic coding task, where agents' capabilities in physical understanding, mathematical formalization and software engineering are put to the test.
    \item  We introduce \textbf{WorldSolver}, comprising 168 tasks derived from physical phenomena in 61 classic computer graphics papers across seven domains, with fixed task-specific scaffolds that isolate solver implementation.
    \item  We propose an evaluation protocol combining execution checks, rendered-simulation assessment, and physics-grounded trajectory verification, benchmarking frontier coding agents under this unified protocol.
\end{itemize}

\section{WorldSolver}
\begin{figure}
    \centering
    \includegraphics[width=1\linewidth]{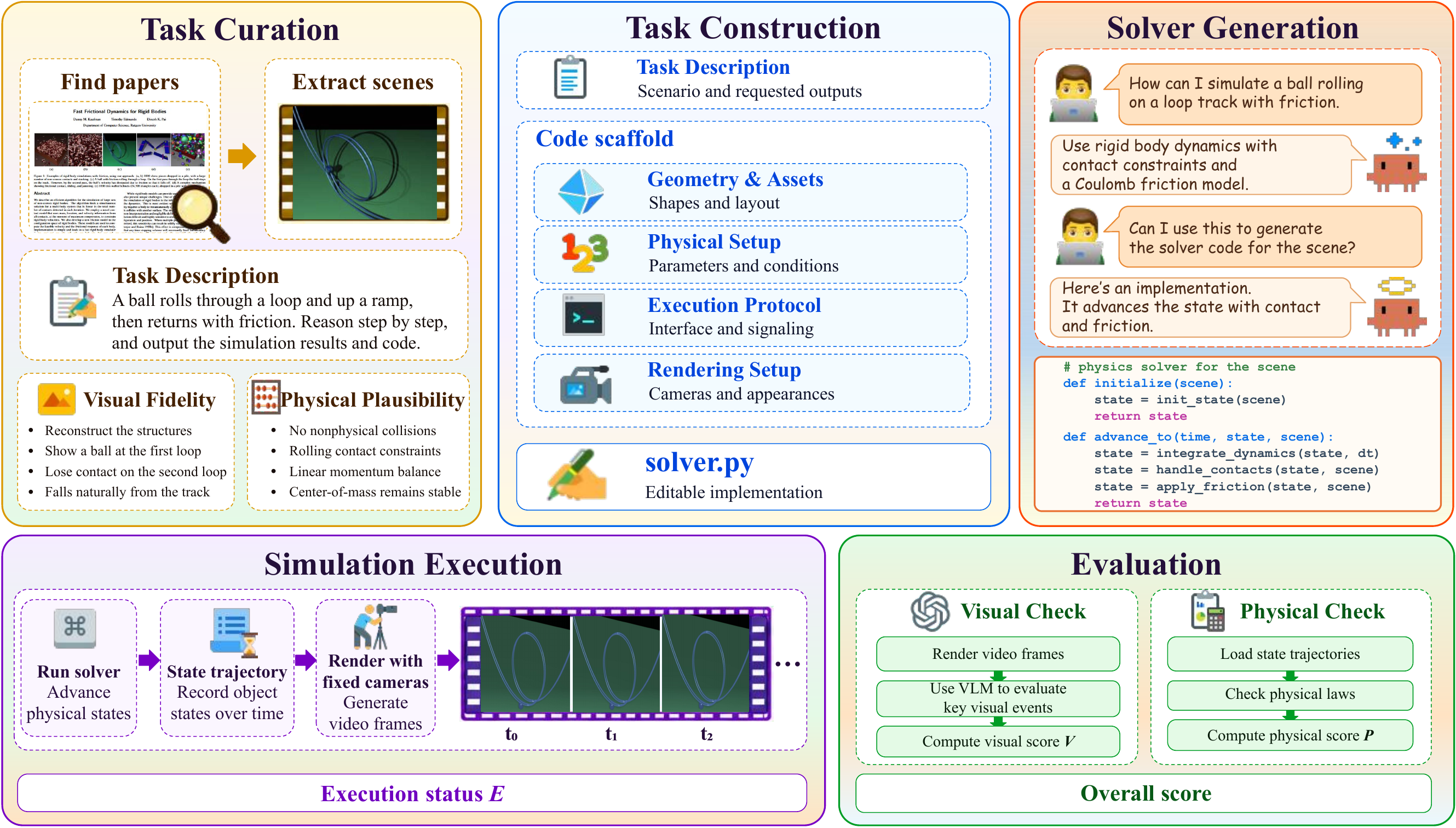}
    \caption{Overview of the WorldSolver workflow.
We curate physical phenomena from classic computer graphics papers and build a simulation task around each one.
Every task supplies a phenomenon description and a code scaffold, and the agent implements the solver.
Running the solver produces state trajectories and rendered videos, which are then assessed by execution, visual, and
  physical checks.}
    \label{fig:pipeline}
\end{figure}

\label{sec:worldsolver}
WorldSolver is a benchmark that evaluates how well coding agents can generate solvers for physical simulation.
Each task is derived from a challenging yet physically realistic phenomenon, curated from classic computer graphics papers spanning seven physical domains.
Overview of the WorldSolver workflow is provided in Figure~\ref{fig:pipeline}.

\subsection{Task Formulation}
\label{sec:task_formulation}
Each task provides a description $d$ of a target physical phenomenon and a fixed code scaffold
\begin{equation}
    \mathcal{S}=(\mathcal{G},\mathcal{P},\mathcal{E},\mathcal{R}),
\end{equation}
where $\mathcal{G}$ specifies scene geometry and assets, $\mathcal{P}$ specifies physical parameters, external inputs, and boundary conditions, $\mathcal{E}$ defines the execution protocol and solver interface, and $\mathcal{R}$ specifies the rendering configuration. The scaffold provides the infrastructure for executing and rendering a simulation.
We then formulate physics solver generation as a task in which an agent $A$ develops a solver $\Phi$ from a description $d$ of a target phenomenon and a fixed code scaffold $\mathcal{S}$:
  \begin{equation}
      \Phi = \mathcal{A}(d, \mathcal{S}).
      \label{eq:solver_generation}
  \end{equation}
  
A solver $\Phi$ defines how the state of a simulated system evolves over time.
Let $\mathbf{s}_t$ denote the physical state at step $t$, and let $\mathbf{s}_0$ be the initial state defined by the scaffold.
Executed under the scaffold's protocol, $\Phi$ advances each state to the next one over a horizon of $T$ steps:
  \begin{equation}
      \mathbf{s}_{t+1} = \Phi(\mathbf{s}_t),
      \qquad t = 0, \dots, T-1.
      \label{eq:state_update}
  \end{equation}
Each state $\mathbf{s}_t$ exports necessary variables for rendering and quantities required by physical checks, such as velocities and contact forces.

Finally, each run produces two outputs for evaluation.
Iterating state $\mathbf{s}_T$ collects a state trajectory $\tau$, and passing the trajectory through the fixed renderer yields a video $v$.
  \begin{equation}
      \tau = (\mathbf{s}_0, \mathbf{s}_1, \dots, \mathbf{s}_T),
      \ v = \mathcal{R}(\tau).
      \label{eq:output}
  \end{equation}

\subsection{Task Construction}
\label{sec:data_construction}

\paragraph{Task Curation.}
We curate our tasks from challenging yet physically realistic phenomena presented in classic computer graphics papers~\citep{guendelman2003nonconvex,li2018animplicit,enright2002animation}.
For each phenomenon, we write a task description together with a set of scoring criteria, the latter withheld from the agent during solver generation.

\begin{figure*}[t]
    \centering
    
    % ================= Left: Task Distribution =================
    \begin{minipage}[c]
    {0.45\textwidth}
        \centering
        \includegraphics[width=\linewidth]
        {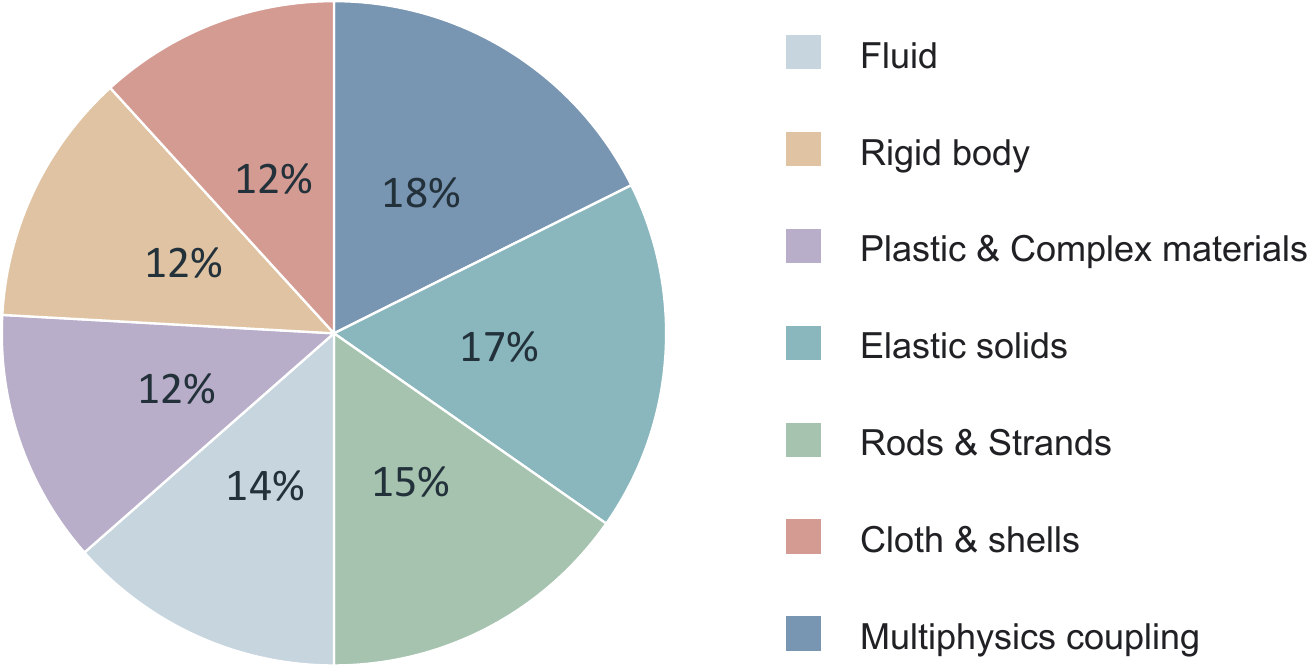}
        \vspace{2mm}
        {\small{(a)}}
    \end{minipage}
    \hfill
    % ================= Right: Domain Cases =================
    \begin{minipage}[c]{0.53\textwidth}
        \centering

        % ---------------- First row ----------------
        \begin{minipage}[t]{0.235\linewidth}
            \centering
            \includegraphics[width=\linewidth]
             {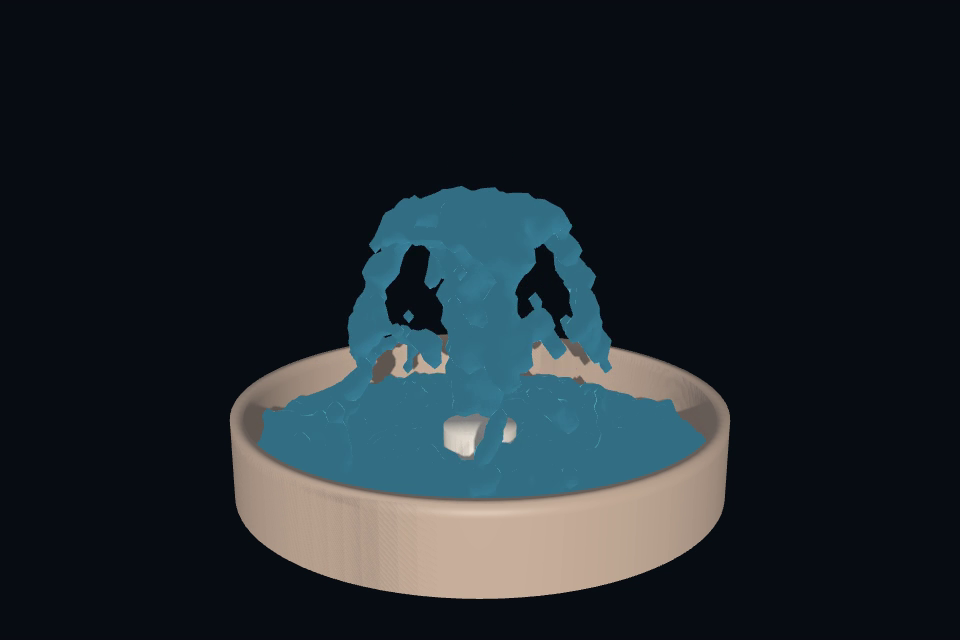}

            \vspace{0.5mm}
            {\tiny Fluid}
        \end{minipage}
        \hfill
        \begin{minipage}[t]{0.235\linewidth}
            \centering
            \includegraphics[width=\linewidth]
            {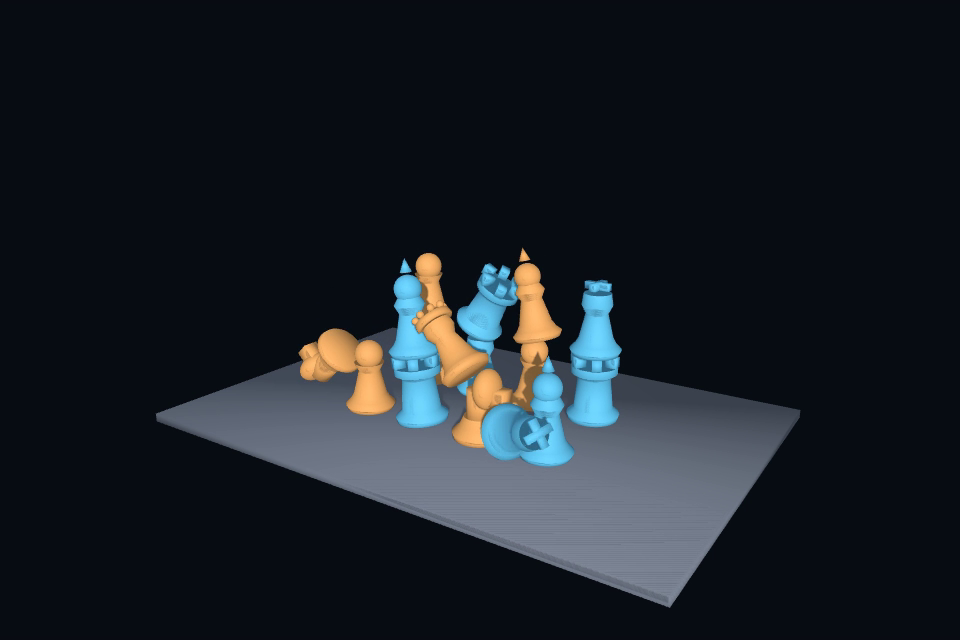}

            \vspace{0.5mm}
            {\tiny Rigid Body}
        \end{minipage}
        \hfill
        \begin{minipage}[t]{0.235\linewidth}
            \centering
            \includegraphics[width=\linewidth]
            {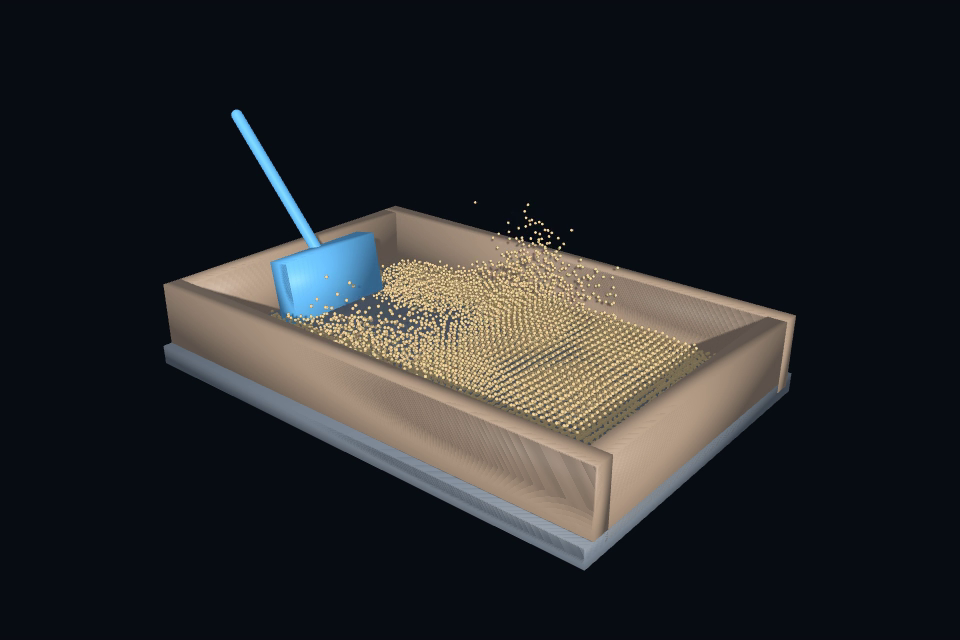}

            \vspace{0.5mm}
            {\tiny Plastic \& Complex\\[-1pt] Materials\par}
        \end{minipage}
        \hfill
        \begin{minipage}[t]{0.235\linewidth}
            \centering
            \includegraphics[width=\linewidth]
            {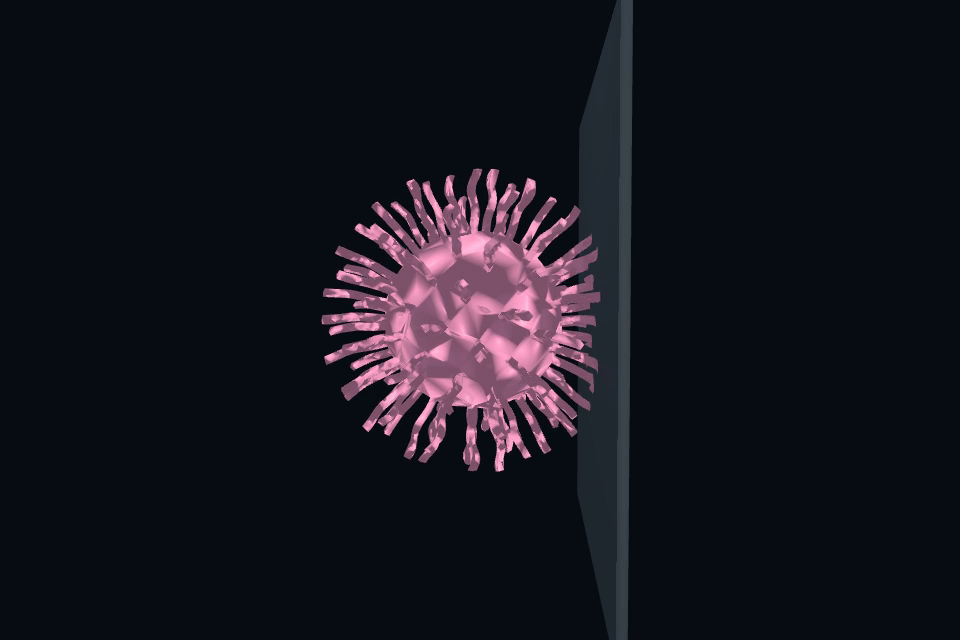}

            \vspace{0.5mm}
            {\tiny Elastic Solids}
        \end{minipage}

        \vspace{2mm}

        % ---------------- Second row ----------------
        \hfill
        \begin{minipage}[t]{0.235\linewidth}
            \centering
            \includegraphics[width=\linewidth]
            {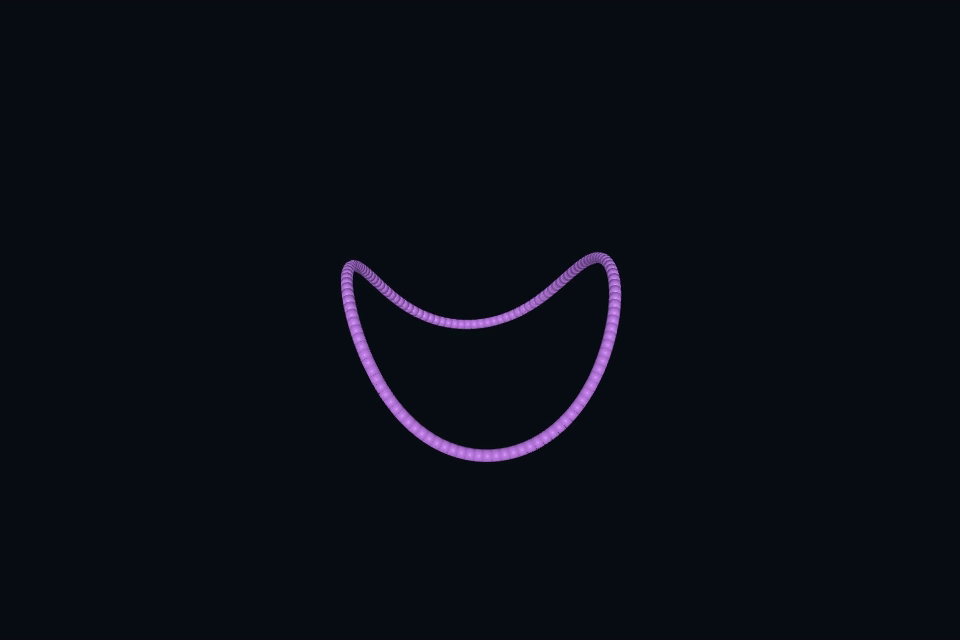}

            \vspace{0.5mm}
            {\tiny Rods \& Strands}
        \end{minipage}
        \hfill
        \begin{minipage}[t]{0.235\linewidth}
            \centering
            \includegraphics[width=\linewidth]
            {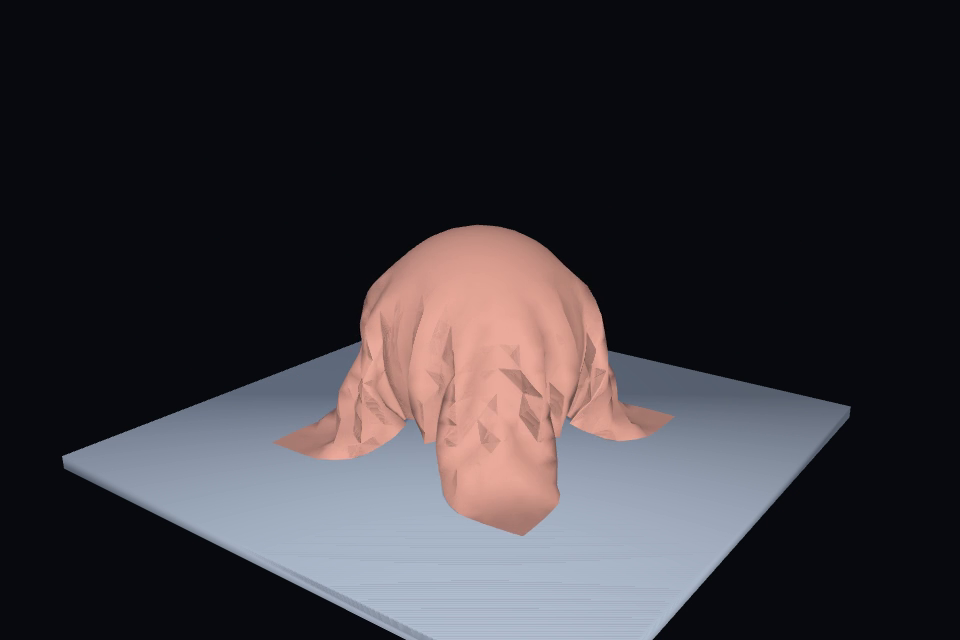}

            \vspace{0.5mm}
            {\tiny Cloth \& Shells}
        \end{minipage}
        \hfill
        \begin{minipage}[t]{0.235\linewidth}
            \centering
            \includegraphics[width=\linewidth]
            {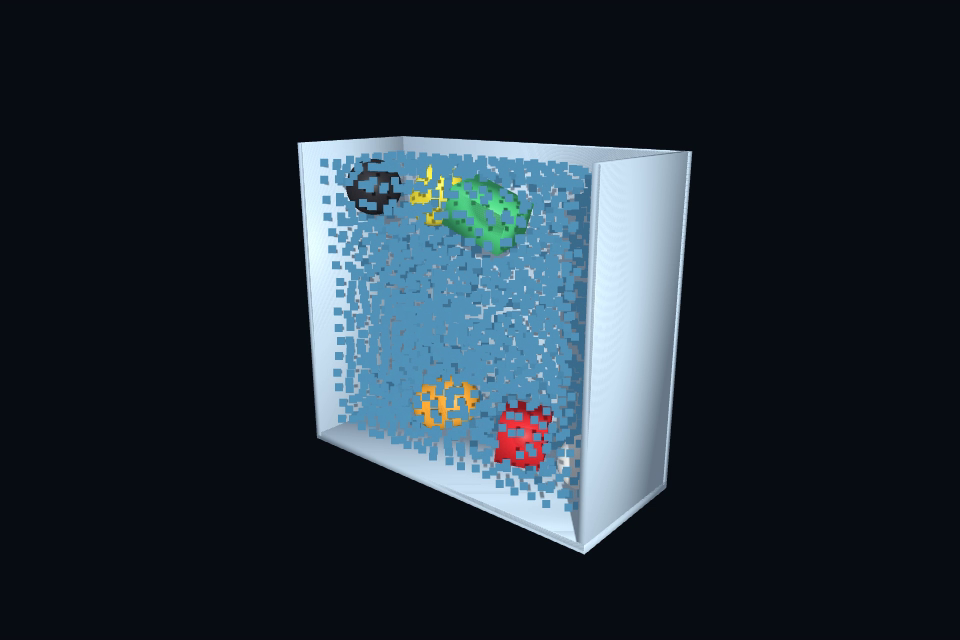}

            \vspace{0.5mm}
            
            {\tiny Multiphysics\\[-1pt] Coupling\par}
        \end{minipage}
        \hfill

        {\small{(b)}}
    \end{minipage}
    
    \caption{Benchmark coverage across seven physics domains. (a) Distribution of the 168 tasks across domains. (b) Representative simulation cases from each domain.}
    \label{fig:domain_cov}
    
\end{figure*}
\begin{table*}[t]
    \centering
    \caption{Comparison with related benchmarks. WorldSolver covers seven physics domains and supports both visual and numerical evaluation of generated solvers. Notably, WorldSolver is the first to use a fixed code scaffold to isolate solver implementation under controlled conditions.}
    \renewcommand{\arraystretch}{1}
    \label{tab:benchmark_comparison}
    \resizebox{\textwidth}{!}{
        \begin{tabular}{lccccc}
            \toprule
            Benchmark
            & \shortstack{Domain\\Coverage}
            & \shortstack{Code\\Scaffold}
            & \shortstack{Solver\\Generation}
            & \shortstack{Visual\\Evaluation}
            & \shortstack{Numerical\\Evaluation} \\
            \midrule

            SimuScene~\citep{wang2026simuscene}
            & Physics (5 domains)
            & $\times$
            & $\times$
            & $\checkmark$
            & $\times$ \\

            PhysCodeBench~\citep{xie2026physcodebench}
            & Physics (3 domains)
            & $\times$
            & $\times$
            & $\checkmark$
            & $\times$ \\

            SimulCost~\citep{cao2026simulcost}
            & Physics (3 domains)
            & $\times$
            & $\times$
            & $\times$
            & $\checkmark$ \\

            CFDLLMBench~\citep{somasekharan2025cfdllmbench}
            & CFD (3 tasks)
            & $\times$
            & $\checkmark$
            & $\times$
            & $\checkmark$ \\

            PDEAgent-Bench~\citep{hang2026pdeagentbench}
            & PDEs (11 families)
            & $\times$
            & $\checkmark$
            & $\times$
            & $\checkmark$ \\

            \midrule
            \textbf{WorldSolver}
            & \textbf{Physics (7 domains)}
            & $\checkmark$
            & $\checkmark$
            & $\checkmark$
            & $\checkmark$ \\
            \bottomrule
        \end{tabular}
    }
    \vspace{2pt}

\end{table*}

\paragraph{Domain Coverage.}
Our benchmark covers seven domains, including fluid~\citep{jiang2015affine,de2015power}, rigid body~\citep{kaufman2005fast,ferguson2021intersection}, plastic and complex materials~\citep{klar2016drucker,stomakhin2013material}, elastic solids~\citep{li2020incremental,irving2007volume}, rods and strands~\citep{bergou2008discrete,bertails2006super}, cloth and shells~\citep{li2021codimensional,bridson2002robust}, and multiphysics coupling~\citep{batty2007afast,guendelman2005coupling}.
Figure~\ref{fig:domain_cov} shows how the tasks distribute across these domains, along with a representative case from each.

\paragraph{Task Specification.}
We turn each description into a fixed code scaffold $\mathcal{S}=(\mathcal{G}, \mathcal{P}, \mathcal{E}, \mathcal{R})$ automatically with manual checks.
The geometry $\mathcal{G}$ and the physical setup $\mathcal{P}$ define what is simulated, from object shapes and arrangements to physical parameters and boundary conditions.
The execution protocol $\mathcal{E}$ and the rendering setup $\mathcal{R}$ define how it is run and observed, fixing the solver interface and trajectory format on one side and the cameras and appearance on the other.
Each agent receives the task prompt $d$ and the same scaffold $\mathcal{S}$ for a given task, with modifications restricted to solver code so that modeling and numerical choices and code engineering are evaluated under controlled conditions.

\subsection{Evaluation Protocol}
\label{sec:evaluation_protocol}

We evaluate submitted solvers along three axes: Execution Checks, Visual Fidelity, and Physical Plausibility. Together, they assess whether a solver runs successfully, reproduces the expected visible phenomena, and respects the relevant physical laws and constraints, with the visual and physical evidence serving as complementary views of simulation quality.

\paragraph{Execution Checks.}
We first run each solver under the code scaffold $\mathcal{S}$, setting the execution score $E=1$ if it completes successfully and produces valid output and $E=0$ otherwise.
For successful executions, the resulting state trajectory $\tau$ and rendered video $v$ are passed to the physical and visual evaluations described below.

\paragraph{Visual Fidelity.}
Visual Fidelity $V$ measures how well the rendered simulation reproduces the task's target phenomena. 
For each task, we specify a set of visual criteria $\mathcal{C}=\{c_i\}_{i=1}^{4}$ across the four shared dimensions shown in Table~\ref{tab:visual_criteria}. 
We use a Vision-Language Model (\text{VLM}) to score each criterion $c_i$ from sampled key frames of the rendered video $v$, and the Visual Fidelity score is their average:

\begin{align}
    V_i = \text{VLM}&(v,c_i) \in [0,1],
    \qquad i=1,\ldots,4,
    \label{eq:visual_criterion}\\
    &V(v;\mathcal{C})
    = \frac{1}{4}\sum_{i=1}^{4} V_i.
    \label{eq:visual_score}
\end{align}
To validate the VLM judge, we evaluate its stability and compare its assessments with those of human experts, as further discussed in Appendix~\ref{app:more_results}.

\begin{table}[t]
    \centering
    \footnotesize
    \setlength{\tabcolsep}{3pt}
    \renewcommand{\arraystretch}{1}
    \caption{Four dimensions of visual fidelity, evaluated using task-specific criteria. The visual fidelity score is the mean score across the four dimensions.}
    \label{tab:visual_criteria}
    \begin{tabular}{@{}p{0.26\linewidth}p{\dimexpr0.74\linewidth-6pt\relax}@{}}
        \toprule
        \textbf{Dimension} & \textbf{Evaluation Focus} \\
        \midrule
        Event Sequence
        & Occurrence and ordering of the key events specified in the task \\
        Interaction Response
        & Visible response of objects and materials to interactions \\
        Signature Phenomenon
        & Reproduction of the characteristic physical phenomenon described in the task \\
        Temporal Coherence
        & Continuity and consistency of motion throughout the simulation \\
        \bottomrule
    \end{tabular}
\end{table}
\paragraph{Physical Plausibility.}
A solver outputs the physical state at every time step, enabling continuous verification beyond the discrete observations used for visual evaluation.
Specifically, Physical Plausibility $P$ verifies whether the generated trajectory $\tau$ respects relevant physical laws and constraints.
For each physical domain, we predefine a set of violation functions $\mathcal{E}_i$ for laws, \textit{such as mass conservation, momentum--impulse consistency, and contact nonpenetration.}
For each case, we define a task-specific rule set $\mathcal{R}={\{r_i\}}_{i=1}^{M}$ based on the relevant physical properties of the simulated phenomenon, with $M$ typically ranging from 3 to 5.
Required output properties are specified in the solver-generation prompt, while the physical laws used for evaluation remain hidden from the agent.
Each rule $r_i$ is evaluated by a calculator $\mathcal{E}_i$ that computes a nonnegative, dimensionless violation error $e_i$:
\begin{equation}
    e_i = \mathcal{E}_i(\tau) \geq 0,
    \qquad i=1,\ldots,M.
    \label{eq:physical_violation}
\end{equation}
The calculator normalizes the raw physical residual and aggregates it over time. 
Thus, $e_i=0$ indicates no measured violation, while larger values indicate stronger violations.
Each violation error $e_i$ is then mapped to a criterion score $p_i\in[0,1]$ using the calculator-specific upper bound, thus higher scores indicate better physical plausibility.
The overall Physical Plausibility score $P\in[0,1]$ is defined as:
\begin{equation}
    P(\tau;\mathcal{R})
    = \frac{1}{M}\sum_{i=1}^{M} p_i.
    \label{eq:physical_score}
\end{equation}

\paragraph{Final Score.}
Visual Fidelity can expose observable failures, but its reliance on discrete key frames may miss errors in the continuous evolution of the system.
Physical Plausibility reveals hidden dynamics from continuous trajectories, but no finite set of metrics can capture every relevant physical rule.
The two therefore serve as complementary evaluation protocols, and we combine their scores for a more complete assessment.
A representative case study illustrating this complementarity is 
provided in the Appendix~\ref{app:case_studies}.
For case $k$, let $E_k\in\{0,1\}$ denote execution success, and let $V_k$ and $P_k$ denote the Visual Fidelity and Physical Plausibility scores.
The case score $s_k$ and final benchmark score are defined as:
\begin{align}
    s_k =
    \begin{cases}
        \dfrac{V_k+P_k}{2}, & E_k=1, \\[6pt]
        0, & E_k=0,
    \end{cases}
    &\qquad k=1,\ldots,K,
    \label{eq:case_score}\\
    \mathrm{Score}
    = \frac{1}{K}\sum_{k=1}^{K}& s_k,
    \qquad K\geq 1.
    \label{eq:final_score}
\end{align}

\section{Experiment}

\definecolor{bestColor}{HTML}{FFF3CC}   
\definecolor{secondColor}{HTML}{E3EFFA} 

\newcommand{\best}[1]{\cellcolor{bestColor}\textbf{#1}}
\newcommand{\second}[1]{\cellcolor{secondColor}#1}

\begin{table}[t]
  \centering
  \caption{Performance across seven physics domains.
Overall denotes the mean score across all cases.
\protect\colorbox{bestColor}{\strut Best} and
\protect\colorbox{secondColor}{\strut second-best}
results are highlighted in each column. All scores are reported on a percentage scale for clarity.}
  \label{tab:main-results}

  \setlength{\tabcolsep}{5pt}
  \renewcommand{\arraystretch}{1.4}
  \resizebox{\columnwidth}{!}{%
    \begin{tabular}{@{}lccccccc!{\hspace{3pt}\vrule\hspace{3pt}}c@{}}
      \toprule
      \makecell[l]{Model}
      & \makecell{Fluid}
      & \makecell{Rigid\\Body}
      & \makecell{Plastic \&\\Complex Materials}
      & \makecell{Elastic\\Solids}
      & \makecell{Rods \&\\Strands}
      & \makecell{Cloth \&\\Shells}
      & \makecell{Multiphysics\\Coupling}
      & \makecell{Overall} \\
      \midrule
      GPT-5.6 Sol~\citep{openaiGpt56}
      & \best{49.6}
      & \best{54.8}
      & 31.2
      & \second{52.6}
      & \best{59.5}
      & \second{35.8}
      & \best{51.5}
      & \best{48.7} \\

      Claude-Opus-5~\citep{anthropicOpus5}
      & \second{48.1}
      & \second{44.4}
      & \best{47.9}
      & \best{61.9}
      & 33.5
      & \best{60.1}
      & \second{34.9}
      & \second{46.7} \\

      Gemini-3.7-Flash~\citep{googleGemini37Flash}
      & 35.8
      & 38.9
      & 26.5
      & 29.5
      & \second{34.0}
      & 9.5
      & 28.4
      & 29.3 \\

      DeepSeek-V4.1-Flash~\citep{deepseekV41Flash}
      & 19.9
      & 31.1
      & 32.1
      & 20.7
      & 31.0
      & 30.0
      & 13.0
      & 24.6 \\

      GLM-5.3~\citep{zaiGlm53}
      & 16.9
      & 30.7
      & \second{33.9}
      & 23.2
      & 31.9
      & 27.9
      & 7.0
      & 23.6 \\

      Qwen-3.8-Max~\citep{alibabaQwen38Max}
      & 19.8
      & 32.0
      & 18.3
      & 18.6
      & 23.9
      & 30.1
      & 19.7
      & 22.7 \\

      Kimi-K2.7-Code~\citep{moonshotKimiK27}
      & 9.7
      & 34.0
      & 20.1
      & 15.4
      & 19.0
      & 12.8
      & 12.3
      & 17.1 \\
      \bottomrule
    \end{tabular}%
  }
\end{table}
\begin{figure}
    \centering
    \includegraphics[width=1\linewidth]{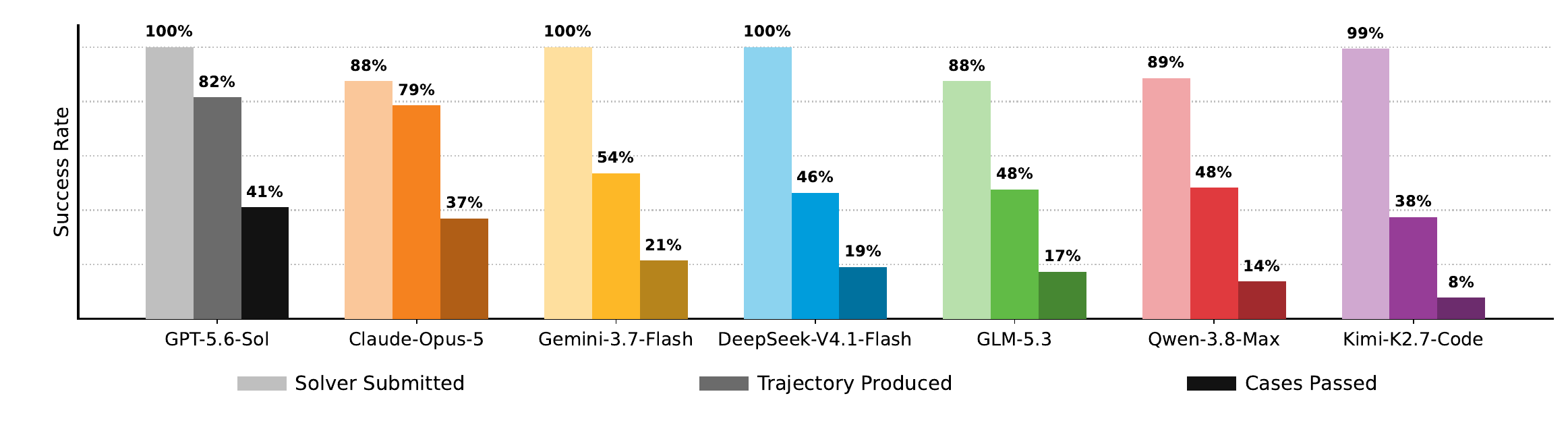}
    \caption{Success rates across three stages of solver generation. 
    Solver Submitted measures successful solver delivery, Trajectory Produced measures successful execution with valid trajectory outputs, and Cases Passed measures the fraction of tasks achieving a final score of at least 60 points.}
    \label{fig:gate}
\end{figure}

\subsection{Experimental Setup}
We evaluate seven frontier models on the WorldSolver benchmark: GPT-5.6-Sol~\citep{openaiGpt56}, Claude-Opus-5~\citep{anthropicOpus5}, Gemini-3.7-Flash~\citep{googleGemini37Flash}, DeepSeek-V4.1-Flash~\citep{deepseekV41Flash}, GLM-5.3~\citep{zaiGlm53}, Qwen-3.8-Max~\citep{alibabaQwen38Max}, and Kimi-K2.7-Code~\citep{moonshotKimiK27}. 
We use Codex CLI as the agent scaffold for GPT-5.6-Sol and Kimi Code for Kimi-K2.7-Code, while all other models are evaluated with Claude Code. 
For each model, we use the reasoning-effort setting of ``high" when available, models without explicit levels are evaluated with thinking mode ``enabled". 
We set a generation time limit of 1,800 seconds for each agent and a runtime limit of 600 seconds for each generated solver. 
We use GPT-5.6-Sol~\citep{openaiGpt56} as the VLM judge and sample 20 key frames from each rendered video for evaluation.
For clarity, all benchmark scores are reported on a 0--100 scale.

\subsection{Main Results}
\paragraph{Overall performance.}
Table~\ref{tab:main-results} shows that physical dynamics solver generation remains challenging for the evaluated agents.
GPT-5.6-Sol achieves the highest overall score of 48.7\%, followed closely by Claude-Opus-5 at 46.7\%.
Gemini-3.7-Flash scores 29.3\%, while DeepSeek-V4.1-Flash, GLM-5.3, Qwen-3.8-Max, and Kimi-K2.7-Code obtain 24.6\%, 23.6\%, 22.7\%, and 17.1\%, respectively.
The performance gaps across models are substantial, yet none of the evaluated agents demonstrates consistently strong solver-generation performance.
We also provide detailed Visual Fidelity and Physical Plausibility results for each agent across all domains in the Appendix~\ref{app:more_results}.

\textbf{Domain-level performance.}
Each agent shows substantial performance variation across different physical domains.
GPT-5.6-Sol performs best on Fluid, Rigid Body, Rods and Strands, and Multiphysics Coupling, whereas Claude-Opus-5 leads on Plastic and Complex Materials, Elastic Solids, and Cloth and Shells.
Although their overall scores differ by only 2 points, their domain-level gaps are much larger: GPT-5.6-Sol outperforms Claude-Opus-5 by 26\% on Rods and Strands, while Claude-Opus-5 leads by 24.3\% on Cloth and Shells.
Several agents also show weaknesses in specific domains, for example, Gemini-3.7-Flash scores only 9.5\% on Cloth and Shells, and GLM-5.3 scores 7.0\% on Multiphysics Coupling.

\begin{figure}[t]
    \centering

    % Left: smoke
    \begin{subfigure}[t]{0.5\linewidth}
        \vspace{0pt}
        \centering
        \includegraphics[width=\linewidth]{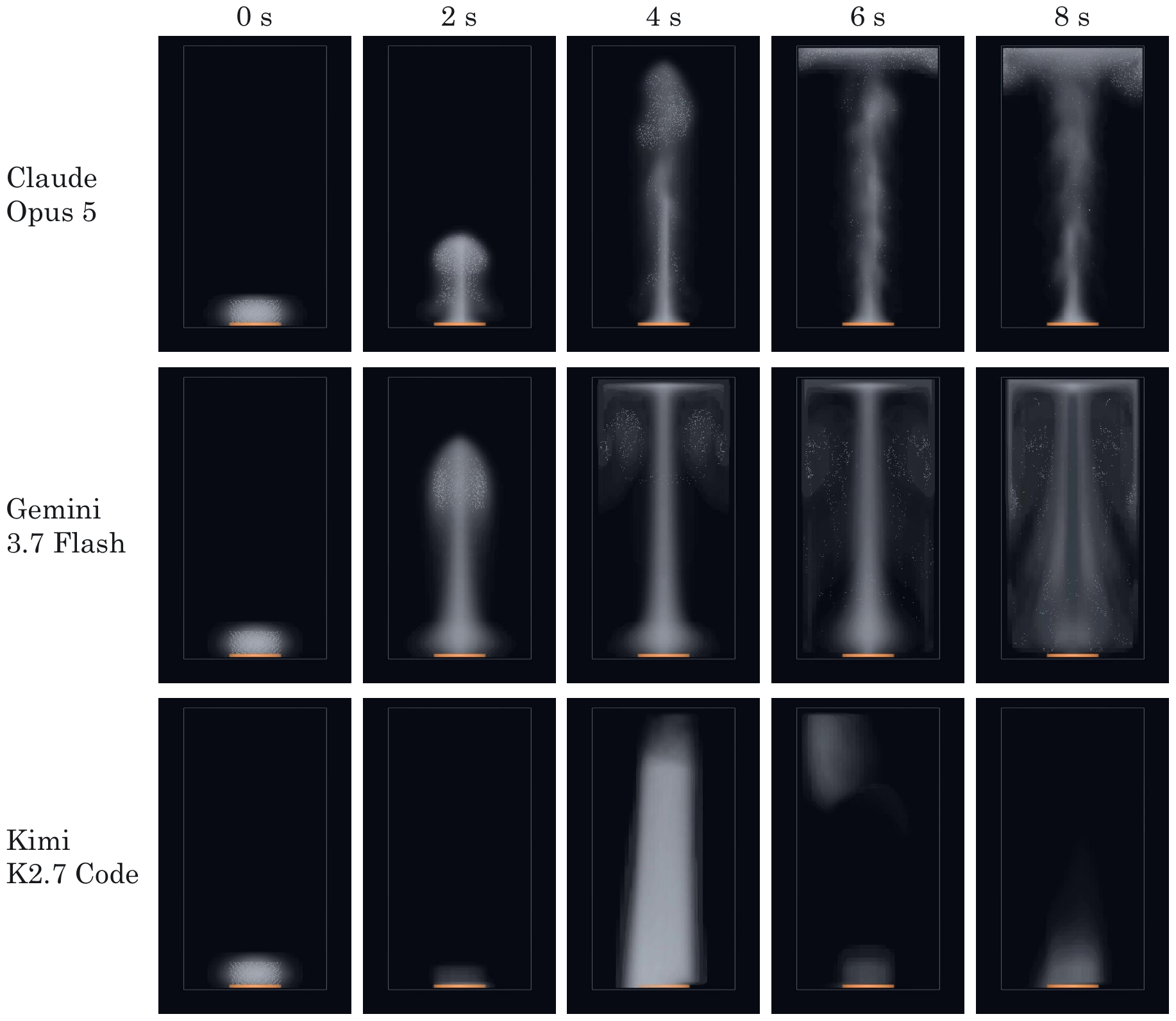}
        \caption{Buoyant smoke}
        \label{fig:showcase_smoke}
    \end{subfigure}
    \hfill
    % Right: dam break and Newton cradle
    \begin{minipage}[t]{0.48\linewidth}
        \vspace{0pt}
        \centering

        \begin{subfigure}[t]{\linewidth}
            \centering
            \includegraphics[width=\linewidth]{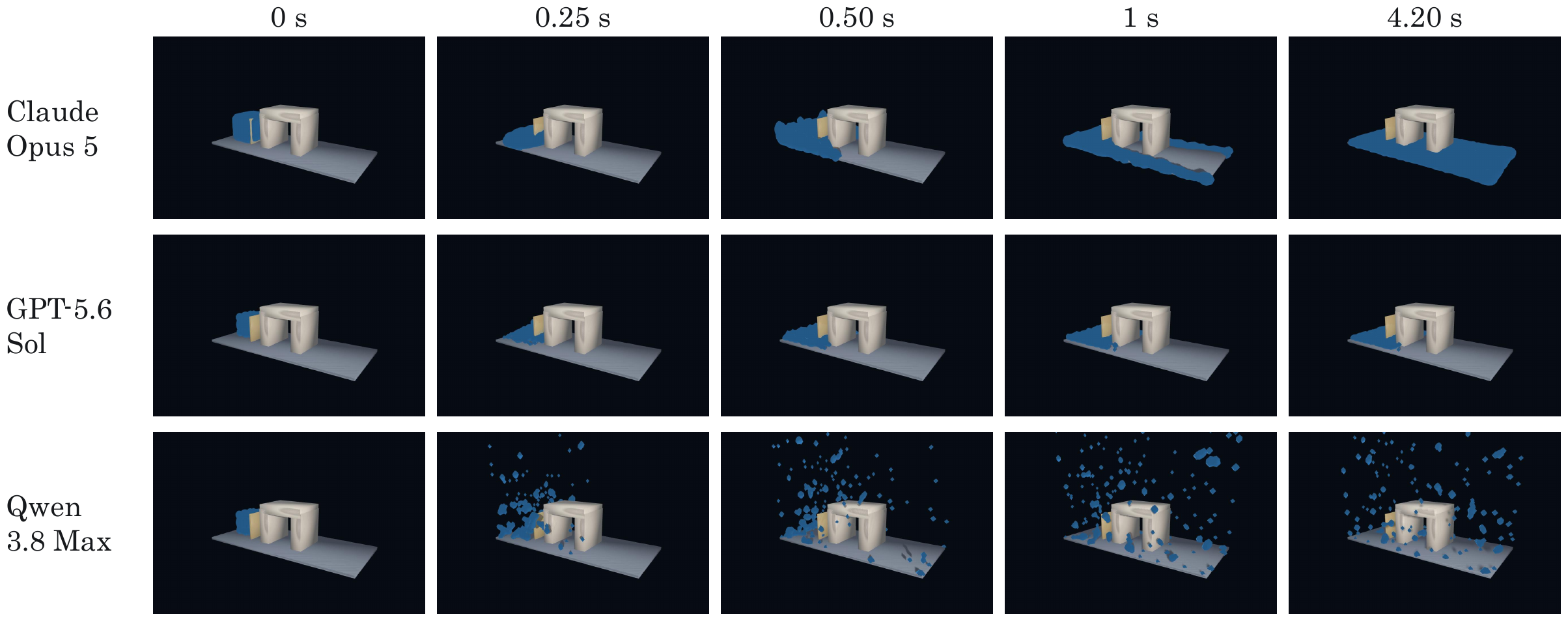}
            \caption{Dam break through an arch}
            \label{fig:showcase_dam}
        \end{subfigure}

        \par\vspace{10pt}

        \begin{subfigure}[t]{\linewidth}
            \centering
            \includegraphics[width=\linewidth]{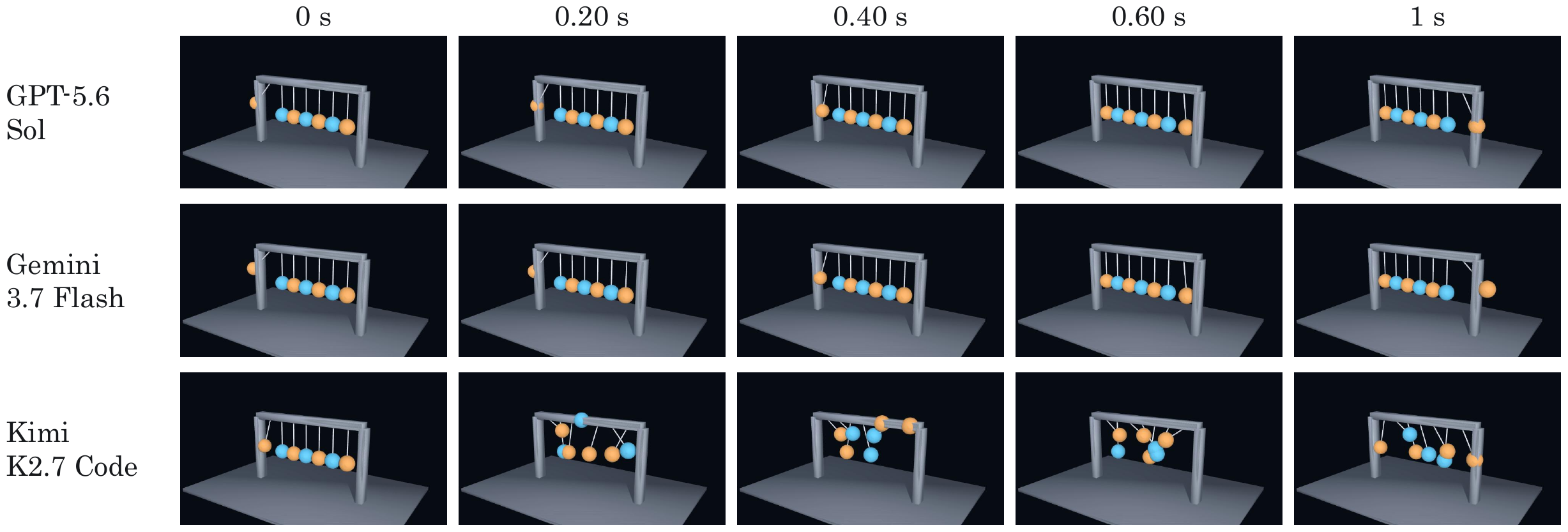}
            \caption{Newton cradle breakaway}
            \label{fig:showcase_cradle}
        \end{subfigure}
    \end{minipage}

    \caption{
        Temporal comparison of selected generated solvers on three
        physical tasks. Within each subfigure, rows identify models
        and columns show matched video timestamps.
        The sequences illustrate differences in plume development,
        flow around an obstacle, and motion following impact.
        Each case uses a fixed crop across models and timestamps.
    }
    \label{fig:solver_showcase}
\end{figure}
\subsection{More Analysis}

\textbf{Success rate across different stages.}
Generating a reliable solver for physical dynamics simulation remains challenging: most agents return a submission, fewer produce valid simulation outputs, and fewer still pass the benchmark threshold.
As shown in Figure~\ref{fig:gate}, \emph{Solver Submitted} denotes the fraction of tasks with a solver returned in time, \emph{Trajectory Produced} the fraction whose solver successfully exports both a trajectory and a rendered video, and \emph{Cases Passed} the fraction achieving a final score of at least 60\%.
We use 60 as a preliminary threshold to illustrate the stage-wise trend, rather than as an established criterion for solver success.
All models submit solvers for most tasks, with rates ranging from 88\% to 100\%.
However, substantially fewer solvers produce valid outputs, from 82\% for GPT-5.6-Sol to 38\% for Kimi-K2.7-Code.
Passing the benchmark threshold is harder still.
GPT-5.6-Sol and Claude-Opus-5 pass 41\% and 37\% of cases, compared with 14\% for Qwen-3.8-Max and 8\% for Kimi-K2.7-Code.
The largest separation appears at the execution stage, where GPT-5.6-Sol and Claude-Opus-5 reach 82\% and 79\%, while the remaining models range from 38\% to 54\%.
This gap is most striking for Kimi-K2.7-Code, which submits a solver for 99\% of tasks but produces valid simulation outputs for only 38\%.

\textbf{Visualization of generated solver dynamics.}
Figure~\ref{fig:solver_showcase} shows the temporal evolution of representative solver across three physical dynamics simulation tasks at matched video timestamps.
In the buoyant smoke example (a), Claude-Opus-5 produces a visually natural rising plume with irregular billowing structures that spread beneath the upper boundary.
Gemini-3.7-Flash develops a more symmetric column with visibly separated particle structures, while Kimi-K2.7-Code shows a transient rise followed by substantial fading, rather than sustained plume development.
In the dam-break example (b), Claude-Opus-5's fluid spreads around the arch and into the downstream region, whereas GPT-5.6-Sol's fluid remains largely upstream over the displayed interval.
Qwen-3.8-Max instead exhibits extensive airborne particle dispersion, without a comparably continuous body of flowing liquid.
In the Newton cradle example (c), GPT-5.6-Sol and Gemini-3.7-Flash show the initially displaced end ball approaching the row and the opposite end ball subsequently swinging outward, with relatively little interior motion.
Kimi-K2.7-Code exhibits large displacements of multiple interior balls, departing from the intended impulse-transfer behavior.

\begin{figure*}[t]
    \centering

    % ================= Left: Mean Epoch =================
    \begin{subfigure}[t]{0.49\textwidth}
        \centering
        \includegraphics[width=\linewidth]{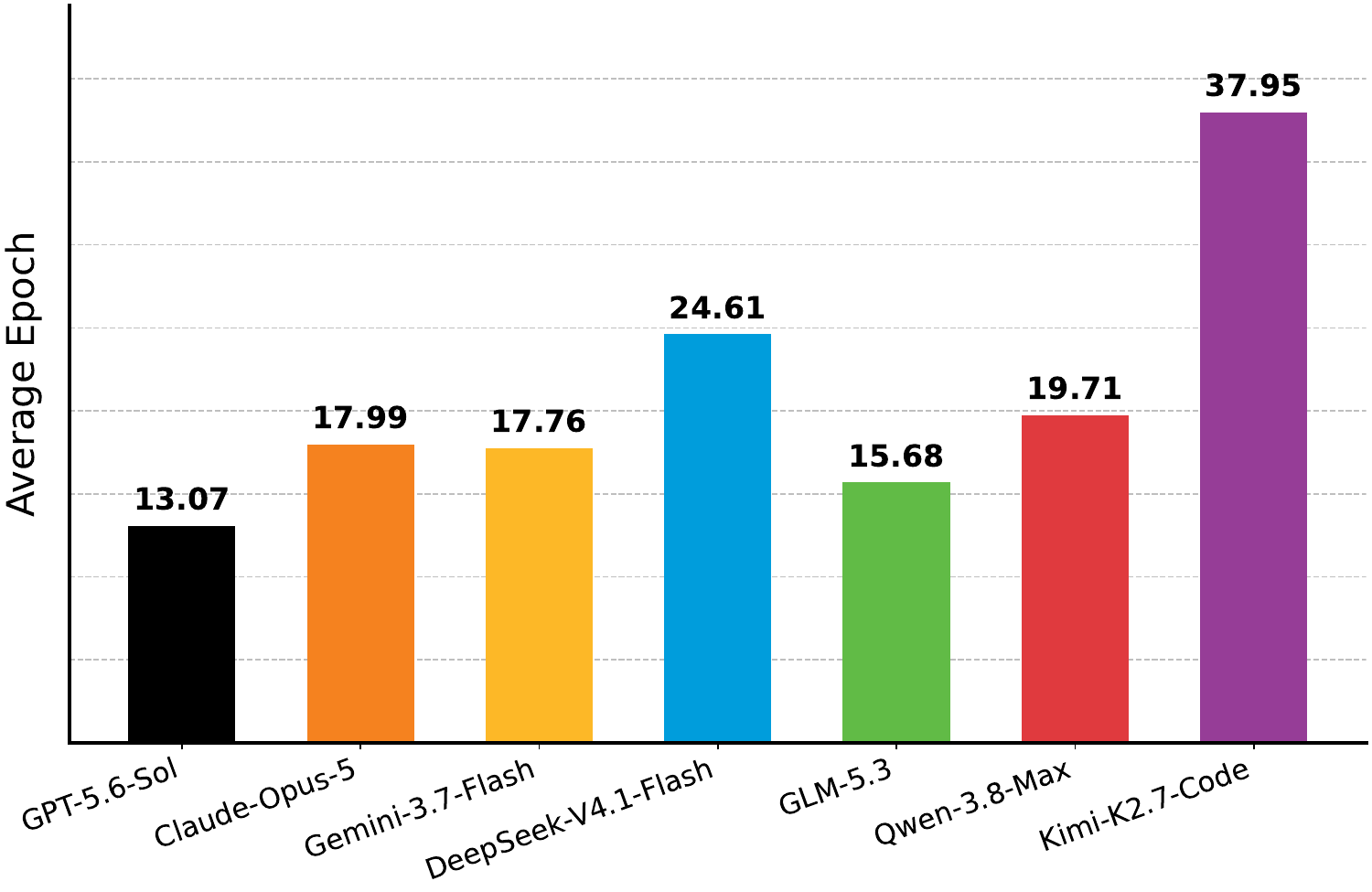}
        \label{fig:mean-epochs}
    \end{subfigure}
    \hfill
    % ================= Right: 2x2 Tool Use =================
    \begin{subfigure}[t]{0.47\textwidth}
        \centering

        \raisebox{2.3cm}[0pt][0pt]{%
        \begin{minipage}{\linewidth}
            \centering

            % ---------- First row ----------
            \begin{subfigure}[t]{0.45\linewidth}
                \centering
                \includegraphics[width=\linewidth]{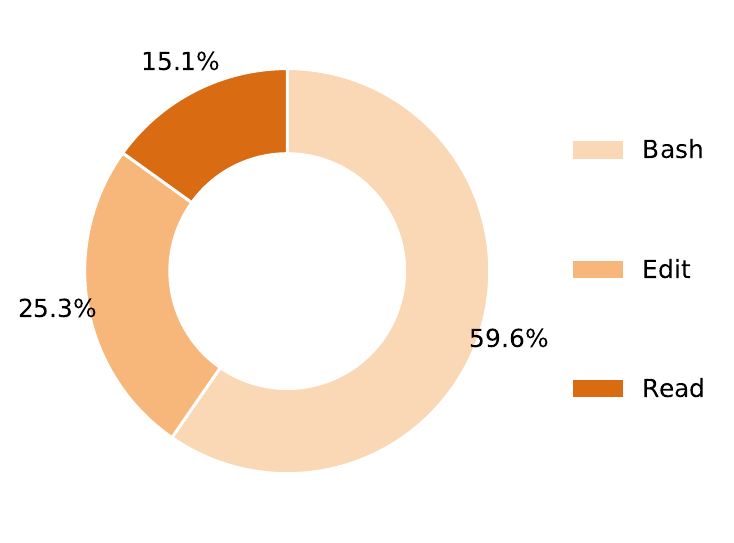}
                \caption{Claude-Opus-5 }
                \label{fig:tool-use-claude}
            \end{subfigure}
            \hspace{0.05\linewidth}
            \begin{subfigure}[t]{0.45\linewidth}
                \centering
                \includegraphics[width=\linewidth]{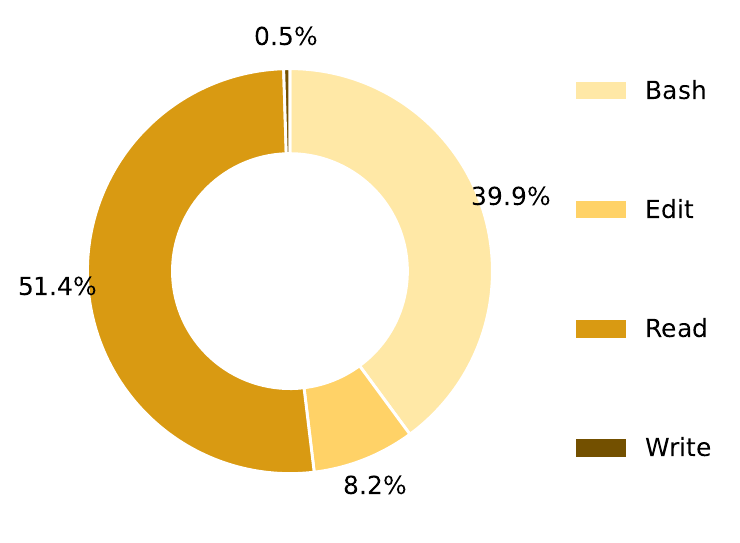}
                \caption{Gemini-3.7-Flash}
                \label{fig:tool-use-gemini}
            \end{subfigure}

            \vspace{0.15em}

            % ---------- Second row ----------
            \begin{subfigure}[t]{0.45\linewidth}
                \centering
                \includegraphics[width=\linewidth]{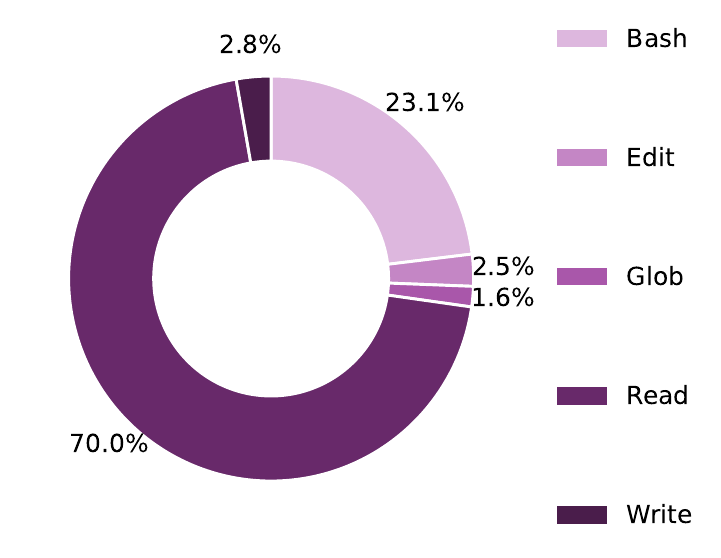}
             
                \caption{Kimi-K2.7-Code}
                \label{fig:tool-use-kimi}
            \end{subfigure}
            \hspace{0.05\linewidth}
            \begin{subfigure}[t]{0.45\linewidth}
                \centering
                \includegraphics[width=\linewidth]{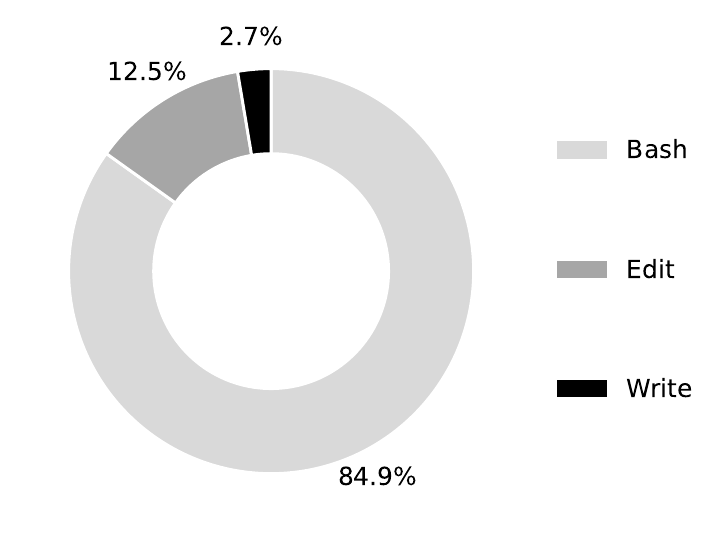}
               
                \caption{GPT-5.6-Sol}
                \label{fig:tool-use-gpt}
            \end{subfigure}

            \vspace{0.2em}

            \label{fig:tool-use-distributions}

        \end{minipage}
        }

    \end{subfigure}

    \caption{Mean epochs and tool-use distributions across the evaluated models. \textit{\textbf{Left:}} Mean model-call epochs, GPT-5.6-Sol averages only 13.07 epochs per task while achieving the highest score. \textit{\textbf{Right:}} Tool-use distributions across $4$ models, Claude-Opus-5 and GPT-5.6-Sol devote more tool-use to editing.}
    \label{fig:epochs-and-tool-use}
\end{figure*}
\begin{figure}[t]
    \centering

\begin{subfigure}[t]{0.48\textwidth}
    \centering
    \includegraphics[width=\linewidth]{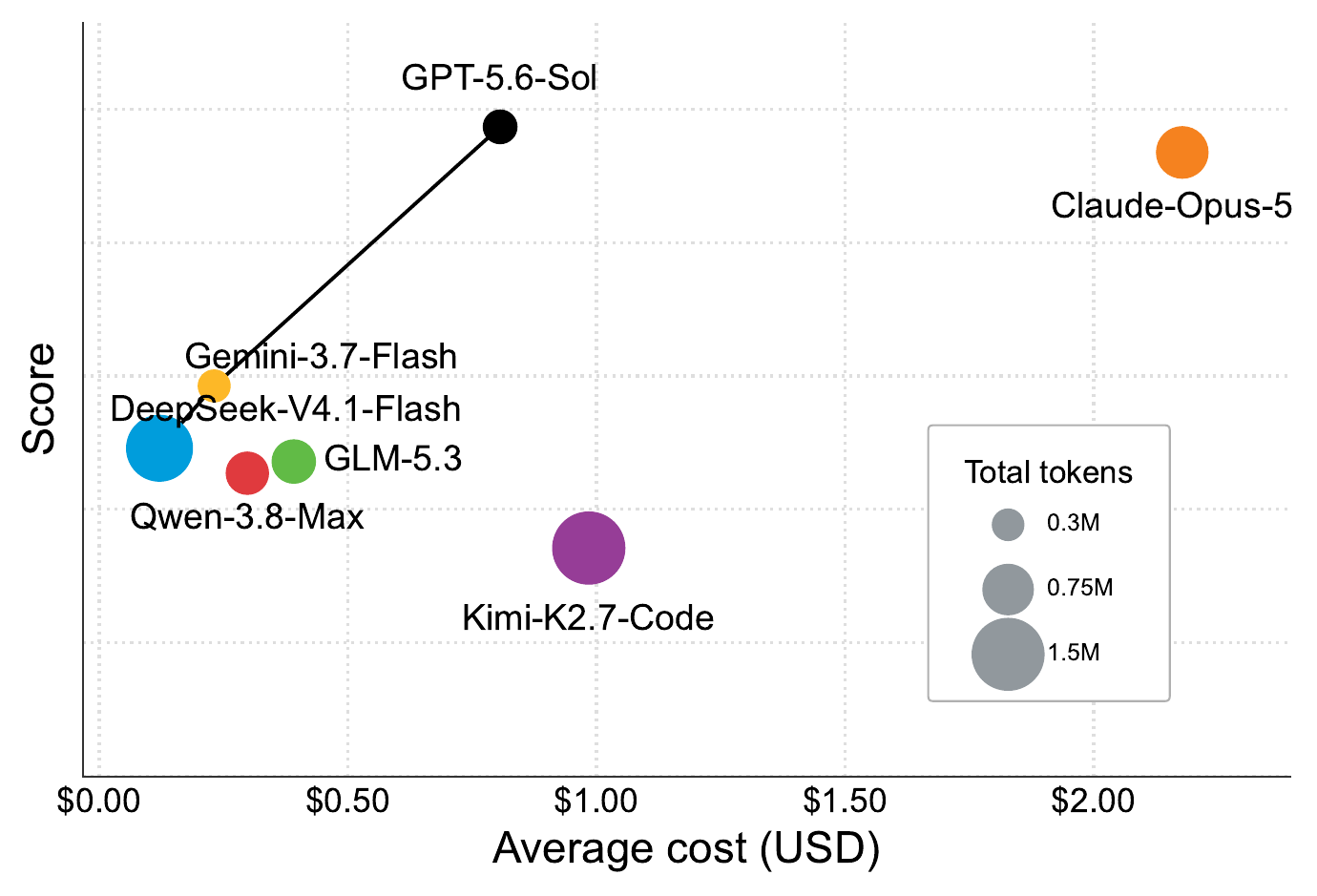}
    \caption{Score versus average API cost.}
    \label{fig:score-vs-usd}
    
\end{subfigure}
\hfill
\begin{subfigure}[t]{0.48\textwidth}
    \centering
    \includegraphics[width=\linewidth]{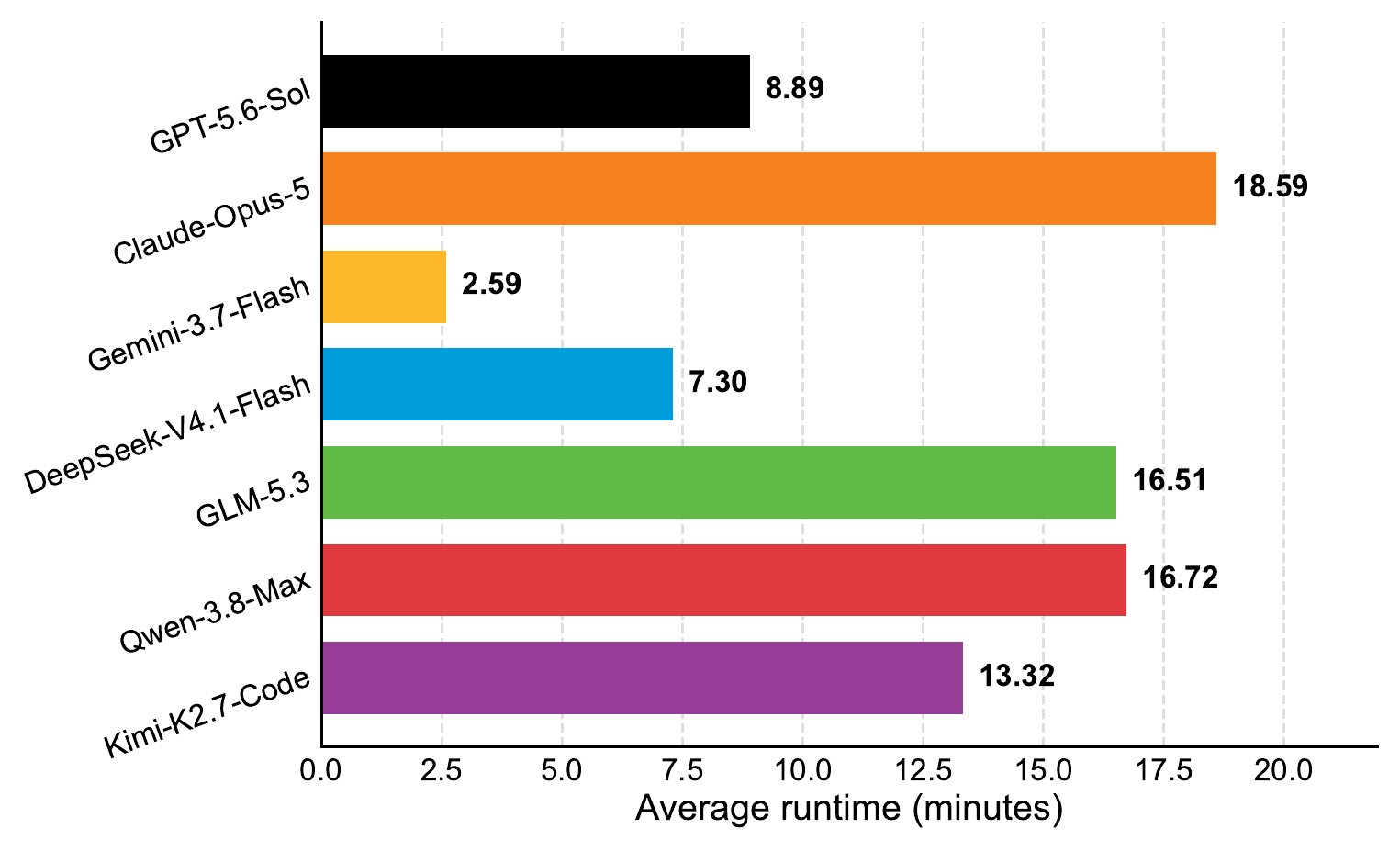}
    \caption{Average wall time.}
    \label{fig:score-vs-minutes}
\end{subfigure}

\caption{Comparison of model score against runtime and API cost. (a) Score versus average API cost, the line represents best cost effectiveness. GPT-5.6-Sol, Gemini-3.7-Flash and DeepSeek-V4.1-Flash demonstrate highest cost effectivenesses. (b) Average agent wall time by model. Gemini-3.7-Flash reports the shortest wall time (2.59 minutes per task on average) while achieving third highest score.}
    \label{fig:score-runtime-cost}
\end{figure}

\vspace{-2mm}
\textbf{Agent behavior patterns in solver generation.}
Figure~\ref{fig:epochs-and-tool-use} shows differences in average epoch counts and tool usage.
GPT-5.6-Sol uses the fewest model-call epochs on average of 13.07, while achieving the highest benchmark score. However, Kimi-K2.7-Code requires 37.95 epochs, nearly three times as many, yet performs substantially worse.
These results show that more model calls do not consistently yield better solvers.
Figure~\ref{fig:epochs-and-tool-use} (a) to (d) shows that Claude-Opus-5 and GPT-5.6 Sol allocate 25.3\% and 12.5\% of their tool calls to Edit, compared with 8.2\% for Gemini-3.7-Flash and 2.5\% for Kimi-K2.7-Code. 
This pattern suggests a potential association between a greater emphasis on code revision and better solver performance.

\textbf{Cost effectiveness}.
Figure~\ref{fig:score-runtime-cost} compares benchmark performance with average API cost and wall time. 
GPT-5.6-Sol achieves the highest score at a lower API cost than Claude-Opus-5, while requiring less than half its average wall time (8.89 versus 18.59 minutes), which is consistent with the mean model-call epochs data. 
Gemini-3.7-Flash offers a cheaper and much faster alternative, with the shortest average wall time of 2.59 minutes, although its score is lower. 
DeepSeek-V4.1-Flash achieves the best cost-effectiveness, but accompanied by high token usage and longer wall time. 
Conversely, Kimi-K2.7-Code incurs a higher API cost and longer wall time than GPT-5.6-Sol despite weaker performance, which is related to its overthinking tendency.

\section{Related Work}
\textbf{Code Generation Benchmarks.}
Code generation aims to translate natural-language specifications into executable programs ~\citep{chen2021evaluating, austin2021program, hendrycks2021measuring, li2022competition}.
As an early milestone, HumanEval~\citep{chen2021evaluating} focuses on generating standalone Python functions and evaluates their correctness through execution.
With the increasing capability of LLM-based agents, recent work has moved beyond function-level generation toward more complex repository-level tasks.
One line of work extends code generation to complex software development tasks, including software engineering ~\citep{jimenez2024swebench,miserendino2025swelancer,yang2024sweagent} and game development ~\citep{jiang2026opengame, chi2026gamedevbench, luo2026gamecraftbench}.
Another line extends code generation to scientific tasks, including implementing scientific ideas ~\citep{huang2024mlagentbench, majumder2024discoverybench, chen2025scienceagentbench} and developing computational tools for scientific computing ~\citep{li2026codepde,hang2026pdeagentbench,somasekharan2025cfdllmbench}.
WorldSolver bridges these two directions by generating physics simulation solvers for described phenomena.
On the one hand, it requires agents to develop and modify physics simulation systems at the repository level.
On the other hand, agents must use physical understanding to select appropriate models and mathematical reasoning to formulate numerical methods.

\textbf{Autonomous Physics Simulation.}
Constructing a physical scene begins with specifying its static configuration, including object geometry, material properties, and initial and boundary conditions. Thanks to the development of LLM models, works in autonomous physics simulation based on LLM agents have emerged recently. SimuScene evaluates code-generated animations through visible behavior~\citep{wang2026simuscene}, while PhysCodeBench,
GS-Agent, and PhysAgent construct dynamic simulations using pre-built physics
engines~\citep{xie2026physcodebench,zhang2026gsagent,lv2026physagent}. Neither setting directly evaluates the full process of selecting physical models and implementing solvers for diverse graphics scenes under fixed experimental conditions. CodePDE, CFD-CodeBench, and PDEAgent-Bench study solver generation for prescribed equations, evaluating solution accuracy, convergence, or computational efficiency~\citep{li2026codepde,somasekharan2025cfdllmbench,hang2026pdeagentbench}. These settings primarily focus on scientific computing, without specific physical scenarios or rendering environments, leaving agents’ physical understanding and mathematical modeling abilities largely unevaluated. WorldSolver instead fixes the scene conditions and requires agents to both select physical models and implement numerical solvers without a pre-built physics engine. Its evaluation combines visible behavior with trajectory-based physical checks across seven domains.

\section{Conclusion}
We introduced WorldSolver, a benchmark for evaluating whether LLM agents can translate physical phenomena description into executable physics solvers. 
It comprises 168 tasks drawn from 61 computer graphics papers across seven physical domains. 
Under fixed code scaffolds, agents implement solvers whose outputs are assessed through execution checks, visual fidelity, and physical plausibility. 
Our experiments show that reliable solver generation remains highly challenging: many agents struggle to submit a runnable solver within the time budget, and even fewer produce solvers that are both visually faithful and physically plausible.
WorldSolver provides a testbed for studying physically grounded computation through agent-generated solvers.
\paragraph{Limitations.}
WorldSolver covers seven physical domains with 168 tasks, but still represents only a subset of physical phenomena.
In the future, we can expand this coverage and explore training methods for physically grounded solver generation.

\bibliography{ref}
\bibliographystyle{iclr2027_conference}

\newpage

\lstdefinestyle{shared}{
    language=Python,
    basicstyle=\small\ttfamily,
    keywordstyle=\color{violet!75!black}\bfseries,
    commentstyle=\color{green!40!black},
    stringstyle=\color{orange!60!black},
    breaklines=true,
    breakatwhitespace=false,
    columns=fullflexible,
    keepspaces=true,
    showstringspaces=false,
    tabsize=4
}

\definecolor{codebg}{RGB}{247,247,247}
\definecolor{codeborder}{RGB}{210,210,210}

\newtcblisting{minimalcode}{
    enhanced,
    listing only,
    breakable,
    colback=codebg,
    colframe=codeborder,
    boxrule=0.5pt,
    arc=1mm,
    left=2mm,
    right=2mm,
    top=1.5mm,
    bottom=1.5mm,
    listing options={style=shared}
}

\definecolor{codeblue}{RGB}{70,110,160}
\definecolor{codebluebg}{RGB}{246,248,251}

\newtcblisting{bluecode}{
    enhanced,
    listing only,
    breakable,
    colback=codebluebg,
    colframe=codebluebg,
    boxrule=0pt,
    borderline west={2.5pt}{0pt}{codeblue},
    sharp corners,
    left=3mm,
    right=2mm,
    top=2mm,
    bottom=2mm,
    listing options={style=shared}
}

\definecolor{titlegray}{RGB}{235,235,235}
\definecolor{bodygray}{RGB}{249,249,249}
\definecolor{bordergray}{RGB}{190,190,190}

\newtcblisting{filecode}[1]{
    enhanced,
    listing only,
    breakable,
    title={\texttt{#1}},
    colback=bodygray,
    colframe=bordergray,
    colbacktitle=titlegray,
    coltitle=black,
    fonttitle=\small\bfseries,
    boxrule=0.6pt,
    arc=1.5mm,
    left=2mm,
    right=2mm,
    top=1.5mm,
    bottom=1.5mm,
    listing options={style=shared}
}

\definecolor{terminalbg}{RGB}{35,38,41}
\definecolor{terminalborder}{RGB}{65,68,72}
\definecolor{terminaltext}{RGB}{235,235,235}
\definecolor{terminaltitle}{RGB}{200,200,200}

\newtcblisting{terminalcode}{
    enhanced,
    listing only,
    breakable,
    title={Terminal},
    colback=terminalbg,
    colframe=terminalborder,
    colbacktitle=terminalbg,
    coltitle=terminaltitle,
    fonttitle=\small\sffamily,
    boxrule=0.6pt,
    arc=2mm,
    left=3mm,
    right=3mm,
    top=1mm,
    bottom=2mm,
    listing options={
        basicstyle=\small\ttfamily\color{terminaltext},
        breaklines=true,
        columns=fullflexible,
        keepspaces=true,
        showstringspaces=false
    }
}

\definecolor{linenumbergray}{RGB}{130,130,130}
\definecolor{codegreenbg}{RGB}{247,250,248}
\definecolor{codegreenborder}{RGB}{185,200,190}

\newtcblisting{numberedcode}[1]{
    enhanced,
    listing only,
    breakable,
    title={\texttt{#1}},
    colback=codegreenbg,
    colframe=codegreenborder,
    colbacktitle=codegreenbg,
    coltitle=black,
    fonttitle=\small\bfseries,
    boxrule=0.6pt,
    arc=1mm,
    left=5mm,
    right=2mm,
    top=1.5mm,
    bottom=1.5mm,
    listing options={
        style=shared,
        numbers=left,
        numberstyle=\scriptsize\color{linenumbergray},
        numbersep=7pt,
        stepnumber=1
    }
}

\newcommand{\styletitle}[2]{%
    \vspace{0.4em}
    {\large\bfseries #1}\hfill{\small\color{gray}#2}\par
    \vspace{0.15em}
}

\appendix

\section{Benchmark Details}

\subsection{Task Curation}
To curate phenomena that are challenging yet physically realistic, we survey computer graphics papers across seven domains and assemble a candidate pool.
Then, we let human experts filter this pool to ensure task quality.
A candidate is retained only if: (1) the phenomenon is visually identifiable; (2) a deterministic scene can be constructed; (3) the observable events can be stated explicitly; (4) the relevant physical quantities can be computed from raw solver outputs.
To keep the benchmark diverse, the experts further merge candidates that share the same objects, geometry, initial state, and target event and differ only in parameter setting. For example, the two-meter-high, ten-body rigid stack settling under gravity appears in its source paper both with and without mass scaling; we keep it as a single task.
The resulting benchmark consists of 168 tasks curated from 61 reference papers.

\subsection{Scaffold Construction}
A code scaffold is the complete simulation setup for a task with the solver removed. 
It defines what the scene contains and how it is observed, but not how the physics is computed: the numerical method is left entirely to the solver.
Every task therefore differs only in its solver, which is the variable we evaluate.
 
To ensure scaffold correctness, we construct each scene through an iterative human-AI refinement procedure. 
A scaffold agent first produces an initial draft from the task description according to a common template. 
Human experts then refine the draft over successive iterations, executing it with a test solver to expose and correct errors in the scene specification. 
The test solver serves a diagnostic purpose only: its trajectories are not treated as physical ground truth, and no solver under evaluation is compared against them. 
A scaffold is accepted once the physical and rendered geometry are consistent, the target event is visible, and the scene and render contracts pass validation. 
Each scaffold adopts the following structure:

 \begin{filecode}{Scaffold Template}
  ID_Name/
      config.yaml
      # Stores the task's public simulation configuration.
      scene.py
      # defines the initial state of the scene
      scene_geometry.yaml
      render_spec.yaml
      # object bindings in the scene.
      scaffold_manifest.json
      run.py
      render.py
      assets/
      solver/
  \end{filecode}

\subsection{Solver Generation}
Each solver agent received the frozen scaffold and a task prompt, and worked in a private workspace.
The prompt supplied four authoritative blocks: the task definition, the public configuration, the scene geometry with its StateFrame bindings, and a task-specific \texttt{Required raw trajectory outputs} block giving the path, shape, and unit of every required field.
These requirements were derived from the physics-metric bindings and the platform's public trajectory-field vocabulary, so the agent exported raw physical state while the platform privately computed the physical plausibility scores.
The agent could read the entire scaffold but modify only \texttt{solver/**}, with no network access, reference material, or files outside its workspace.
Existing simulation engines were barred, so the physical model, discretization, and numerical scheme had to be derived from base scientific libraries.
Specifically, the solver implemented three methods:
 \begin{filecode}{Solver.py}
  class BenchmarkSolver:
      def initialize(self, scene, seed: int) -> None:
          raise NotImplementedError("Solver Agent must implement initialize()")

      def advance_to(self, time: float):
          raise NotImplementedError("Solver Agent must implement advance_to()")

      def diagnostics(self) -> dict[str, object]:
          return {}
  \end{filecode}
The solver was required to perform physical time stepping and to respect the frozen configuration, geometry, collisions, and state bindings. 
Hard-coded animations, keyframes, precomputed trajectories, and scene-specific motion tables were prohibited. 
Left unconstrained, a solver generation agent tends to keep designing a full engine, or to iterate on parameters indefinitely, and may submit nothing at all.
Our prompt therefore pushed it toward an early deliverable: save a complete physically-based baseline well before the allowance is spent, verify it with a compile check, and make at most two focused correction passes for concrete syntax, interface, required-field, or obvious implementation errors found by inspection.
No token or tool-call limit was stated.

\subsection{Task Outputs}
After submission the solver source was frozen, and the platform executed it, recorded the trajectory, rendered the video, and evaluated the result with no further source changes.

  \begin{filecode}{trajectory.h5}
  / metadata_json={"scene_id": "017_ten_body_stack", "seed": 17017}

  /time  float64 (301,)
      [0.0, 0.02, 0.04, ...]

  /frames/00000000/
      position  float64 (10, 3)
          [[ 0.015, 0.0, 0.150],
           [-0.012, 0.0, 0.355],
           [ 0.010, 0.0, 0.560], ...]
      velocity  float64 (10, 3)
          [[0.0, 0.0, 0.0], [0.0, 0.0, 0.0], [0.0, 0.0, 0.0], ...]
      orientation  float64 (10, 4)
          [[0.0, 0.0, 0.0, 1.0], [0.0, 0.0, 0.0, 1.0], [0.0, 0.0, 0.0, 1.0], ...]
      angular_velocity  float64 (10, 3)
          [[0.0, 0.0, 0.0], [0.0, 0.0, 0.0], [0.0, 0.0, 0.0], ...]
      mass  float64 (10,)
          [1.0, 1.0, 1.0, ...]
      body_id  int64 (10,)
          [1, 2, 3, ...]
      scalars/  attrs: contact_length_scale=0.2
      attributes/
      fields/
          contact_signed_distance  float64 (10,)
              [0.050, 0.005, 0.005, ...]
          external_force  float64 (10, 3)
              [[0.0, 0.0, -9.81], [0.0, 0.0, -9.81], [0.0, 0.0, -9.81], ...]
          interaction_force  float64 (10, 3)
              [[0.0, 0.0, 0.0], [0.0, 0.0, 0.0], [0.0, 0.0, 0.0], ...]

  /frames/00000001/  ...
  /frames/00000300/  ...
  \end{filecode}
  
\subsection{Execution Environment}
We separate the agent-generation environment from post-submission execution and evaluation. 
Agents run in disposable Debian Bookworm containers with only the assigned workspace mounted, holding NumPy, SciPy, Numba, h5py, PyYAML, and a bundled FFmpeg. 
The evaluation package is removed from these images, so the scoring metrics are hidden from the agents.
Frozen submissions are then executed, rendered, and evaluated host-side by the platform.
Rendering is performed by offscreen OpenGL rasterization through pyrender and PyOpenGL, trimesh for geometry processing, and ImageIO with FFmpeg for video encoding.

\section{Evaluation Details}
\label{app:evaluation_details}

\subsection{Execution Checks}
The execution check determines whether a submitted solver can run successfully and generate the required outputs. 
A case fails this check if it cannot export an evaluable trajectory and rendered video. 
Common failures include incomplete implementations, interface violations, runtime errors, numerical instability, and time exhaustion.

\subsection{Visual Fidelity Evaluation}
Our visual fidelity evaluation is divided into four parts: task-specific
visual criteria construction, uniform frame sampling, VLM-based judging,
and repeated score aggregation.

\textbf{Part 1: Task-specific visual criteria.}
For every task, we construct four visual criteria from the target physical
phenomenon and its required events. 
The criteria follow four shared
dimensions: \texttt{event\_sequence}, \texttt{interaction\_response},
\texttt{signature\_phenomenon}, and \texttt{temporal\_coherence}.
As an example, the Cube Drop Water task considers a rigid cube entering
water, producing a splash and wake, and subsequently moving according to
its buoyant response. 
Its task-specific visual criteria are shown below.

\begin{filecode}{metric.yaml}
visual:
  - id: visible_event_onset
    dimension: event_sequence
    criterion: The rigid cube visibly falls toward and enters
      the water surface.

  - id: visible_interaction_response
    dimension: interaction_response
    criterion: Water entry produces a visible splash and a
      depressed free surface around the cube.

  - id: visible_signature_outcome
    dimension: signature_phenomenon
    criterion: During subsequent motion, a wake forms and the
      cube continues to sink, rise, or float according to its
      buoyant response.

  - id: case_visual_coherence
    dimension: temporal_coherence
    criterion: The coupled motion remains continuous without
      cube-water interpenetration, surface sticking,
      teleportation, or a fluid response detached from the
      cube trajectory.
\end{filecode}

\textbf{Part 2: Uniform frame sampling.}
We uniformly sample 20 frames over the full timeline of each rendered MP4.
For a video containing \(N\) decodable frames, the sampled frame indices are
\[
i_k =
\mathrm{round}\!\left(\frac{k(N-1)}{19}\right),
\qquad k=0,\ldots,19.
\]
This sampling includes the first and last frames. 
The sampled frames are kept in chronological order, resized so that each side is at most 768 pixels, and encoded as JPEG images. 
Each image is labeled with its temporal position, such as \texttt{(frame 1/20)}.

\textbf{Part 3: VLM-based judging.}
We use \texttt{GPT-5.6-Sol} as the visual judge. 
The 20 ordered frames and the task-specific visual criteria are supplied in the same request. 
The judge assigns each criterion a score between 0 and 1 and provides a short
evidence statement. 
The key prompt content is shown below.
\begin{figure}[h]
    \centering
    \includegraphics[width=\linewidth]{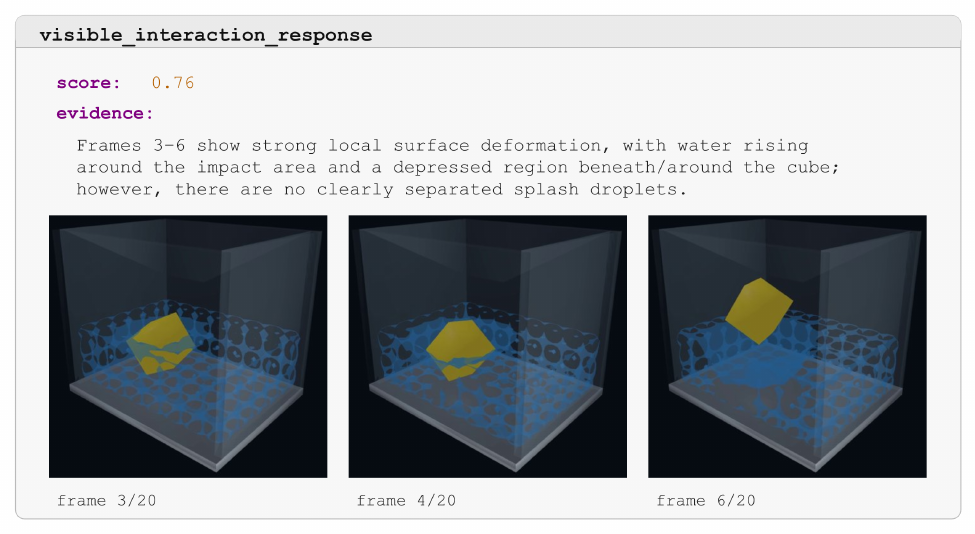}
    \caption{An example of VLM output for the interaction-response criterion on case ``Cube Drop Water''. }
    \label{fig:VLM}
\end{figure}

\begin{filecode}{VLM Judge Prompt}
System:
You are a strict visual benchmark judge. Inspect the attached
frames in chronological reading order. Judge only visible
evidence. Return JSON only.

User:
The image blocks above are video frames in chronological order,
each labeled (frame i/N).

Evaluate every rubric item below.

For each item return criterion_id, score (number from 0 to 1,
where 1 is best and 0 is worst), and a concise evidence
statement. Penalize missing tank/subject visibility,
discontinuity, and absent required events.

Rubric:
[task-specific visual criteria]

Required JSON shape:
{"criteria": [
  {
    "criterion_id": "...",
    "score": 0.0,
    "evidence": "..."
  }
]}
\end{filecode}

Figure~\ref{fig:VLM} shows an example of how VLM outputs score based on the evidence together. The GPT-5.6-Sol solver output shows a partially immersed cube with strong surface deformation but the splash is not clear enough.

\textbf{Part 4: Repeated evaluation and score aggregation.}
We evaluate the same key frames five times using the same prompt. 
For each criterion, we average the five returned scores. 
The final visual score is the average of the four averaged criterion scores. 
The visual fidelity score for the Cube Drop Water task is shown below.

\begin{filecode}{evaluation.json}
{
  "criteria": {
    "visual": [
      {
        "criterion_id": "visible_event_onset",
        "score": 0.978,
        "details": {
          "attempt_count": 5,
          "frame_count": 20
        }
      },
      {
        "criterion_id": "visible_interaction_response",
        "score": 0.796,
        "details": {
          "attempt_count": 5,
          "frame_count": 20
        }
      },
      {
        "criterion_id": "visible_signature_outcome",
        "score": 0.846,
        "details": {
          "attempt_count": 5,
          "frame_count": 20
        }
      },
      {
        "criterion_id": "case_visual_coherence",
        "score": 0.772,
        "details": {
          "attempt_count": 5,
          "frame_count": 20
        }
      }
    ]
  },
  "visual_score": 0.848
}
\end{filecode}
\textbf{Example interpretation.}
The cube entry is clearly visible and therefore receives the highest
criterion score. The splash, subsequent buoyant motion, and overall
temporal coherence are also captured, although the cube--water interface
is less clear in some frames. The criterion-level results expose visual
weaknesses that would be hidden by reporting only the final score.

\subsection{Physical Plausibility Evaluation}

The physical plausibility evaluation is divided into three parts:
task-specific criterion binding, deterministic calculator execution, and
result inspection.

\textbf{Part 1: Task-specific physical criteria and calculator bindings.}
Physical criteria are selected according to the domain and target phenomenon
of each task. Each criterion contains a \texttt{dimension} that groups related
physical properties, while the \texttt{calculator} specifies the numerical
check to execute. As an example, we use the \texttt{GPT-5.6-Sol} result for
the Billiard Impulse Chain task. The task evaluates
nonpenetrating contact, linear momentum balance, center-of-mass consistency,
and internal interaction-force balance. We fully expand the first binding
below and abbreviate the remaining entries.

\begin{filecode}{metric.yaml}
physics:
  - id: nonpenetrating_contact
    dimension: contact_and_interface_conditions
    numeric_binding:
      calculator: contact.penetration
      inputs:
        signed_distance: fields.contact_signed_distance
        length_scale: scalars.contact_length_scale
      parameters: {}
      reduction:
        type: maximum

  # Remaining entries follow the same structure and are abbreviated.
  - id: linear_momentum_balance
  - id: center_of_mass_consistency
  - id: internal_interaction_force_balance
\end{filecode}

The names on the left side of the \texttt{inputs} block are calculator
arguments, and the values on the right are paths to the corresponding
trajectory fields. The \texttt{parameters} block provides additional fixed
settings, while \texttt{reduction} specifies how the per-frame errors are
combined.

\textbf{Part 2: Deterministic calculator execution.}
The evaluator resolves the input paths, executes the selected calculator
over the trajectory, and reduces its per-frame errors to one normalized
error. The central execution logic is:

\begin{filecode}{evaluator.py}
binding = criterion["numeric_binding"]
calculator = CALCULATORS[binding["calculator"]]

result = calculator.compute(
    frames,
    binding["inputs"],
    binding.get("parameters", {}),
)

normalized_error = reduce_values(
    result.normalized_errors,
    binding["reduction"],
)
\end{filecode}

The contact calculator measures the deepest ball--ball or ball--support
penetration relative to the contact length scale. The momentum calculator
compares the change in total system momentum with the exported external
forces. The center-of-mass calculator compares each ball's observed
displacement with the displacement obtained by integrating its velocity.
The interaction-force calculator checks whether the exported internal
contact forces cancel in aggregate.

\textbf{Part 3: Example results.}
For the Billiard Impulse Chain example, we fully expand the first result.
The remaining entries follow the same structure and retain only the fields
needed for identification and comparison.

\begin{filecode}{evaluation.json}
{
  "task_id": "012_billiard_impulse_chain",
  "solver_model": "gpt-5.6-sol",
  "physics": [
    {
      "criterion_id": "nonpenetrating_contact",
      "calculator": "contact.penetration",
      "reduction": {
        "type": "maximum"
      },
      "normalized_error": 0.0,
      "details": {
        "sample_count": 181,
        "worst_frame_index": 0
      }
    },

    // Remaining entries follow the same structure and are abbreviated.
    {
      "criterion_id": "linear_momentum_balance",
      "normalized_error": 1.0
    },
    {
      "criterion_id": "center_of_mass_consistency",
      "normalized_error": 0.101274
    },
    {
      "criterion_id": "internal_interaction_force_balance",
      "normalized_error": 0.0
    }
  ],
  "physical_score": 73.734
}
\end{filecode}

The result contains no detected penetration, and the exported internal
contact forces cancel correctly. The center-of-mass error is 0.101, with a
maximum position residual of approximately \(0.0149\) meters. However, the
total momentum change is inconsistent with the exported external-force
record, producing a momentum-balance error of \(1.0\). The simulation
therefore receives a physical score of \(73.734\), compared with a visual
score of \(99.5\). It reproduces the expected motion convincingly in the
rendered video, while the trajectory reveals a momentum inconsistency that
is difficult to detect visually.

\section{Case Studies}
\label{app:case_studies}
To examine the complementarity of Visual Fidelity and Physical Plausibility, we present a fine-grained case study that illustrates the behavior of our metric.
Specifically, we group the results into three categories: Visually Correct but Physically Incorrect, Physically Correct but Visually Inaccurate, and Metrics in Agreement.
Each category is illustrated with four pairs of cases, accompanied by a clarification that traces the judgment back to its root cause.
Together, these examples show that neither score is sufficient in isolation: visual fidelity evaluates whether the expected phenomenon is visibly reproduced, whereas physical plausibility tests whether the generated dynamics satisfy the selected quantitative constraints.
\subsection{Right Appearance, Wrong Physics}
\begin{figure*}[h]
    \centering
    \includegraphics[width=0.8\textwidth]{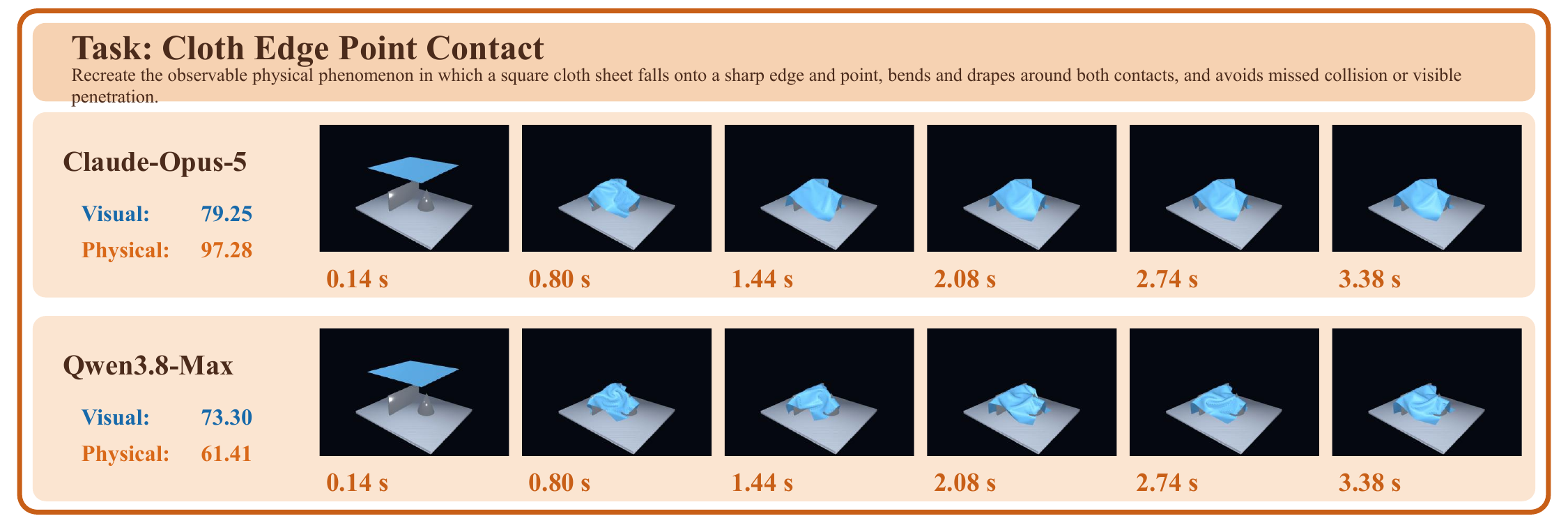}
    \caption{Case study results for the cloth edge--point contact task.}
    \label{fig:physical_B031}
\end{figure*}

Figure~\ref{fig:physical_B031} simulates a square cloth sheet falling onto a sharp edge and an isolated point. Claude-Opus-5 and Qwen-3.8-Max obtain relatively similar visual scores, because both videos show the expected falling cloth, obstacle interaction, and folded resting configuration. 
However, the physical evaluation separates the two solutions much more clearly. 
The main differences arise from the \texttt{orientation\_preserving\_deformation} and \texttt{nonpenetrating\_contact} criteria. 
Qwen-3.8-Max's simulation remains visually recognizable as draping cloth, but its physical trajectory contains extensive element inversion and obstacle interpenetration that are difficult to diagnose from rendered frames alone.

\begin{figure*}[h]
    \centering
    \includegraphics[width=0.8\textwidth]{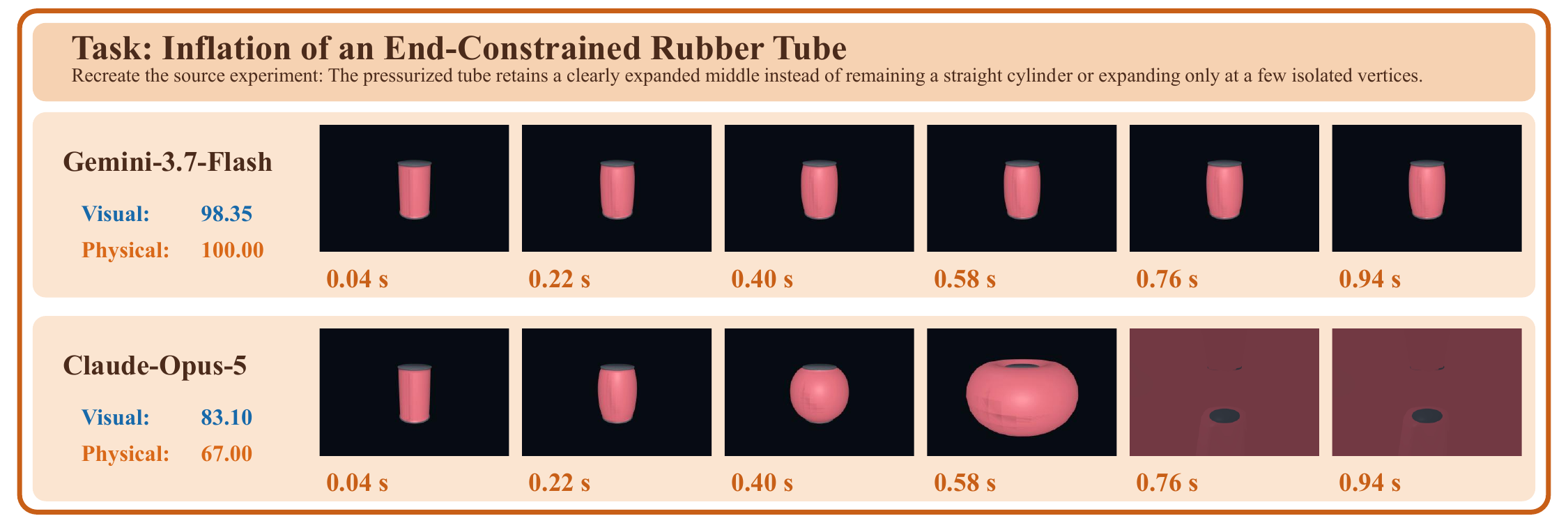}
   \caption{Case study results for the pressurized rubber tube task.}
    \label{fig:physical_B058}
\end{figure*}

As shown in Figure~\ref{fig:physical_B058}, Gemini-3.7-Flash and Claude-Opus-5 both receive relatively high visual scores because both videos show the tube expanding outward while retaining its constrained ends. 
However, the physical evaluation clearly favors Gemini-3.7-Flash. 
The main difference arises from the \texttt{quasistatic\_nodal\_force\_balance} criteria.
Claude-Opus-5 produces a plausible inflated silhouette, but its unconstrained material nodes have large residual forces, indicating that the displayed configuration is not close to mechanical equilibrium.

\begin{figure*}[h]
    \centering
    \includegraphics[width=0.8\textwidth]{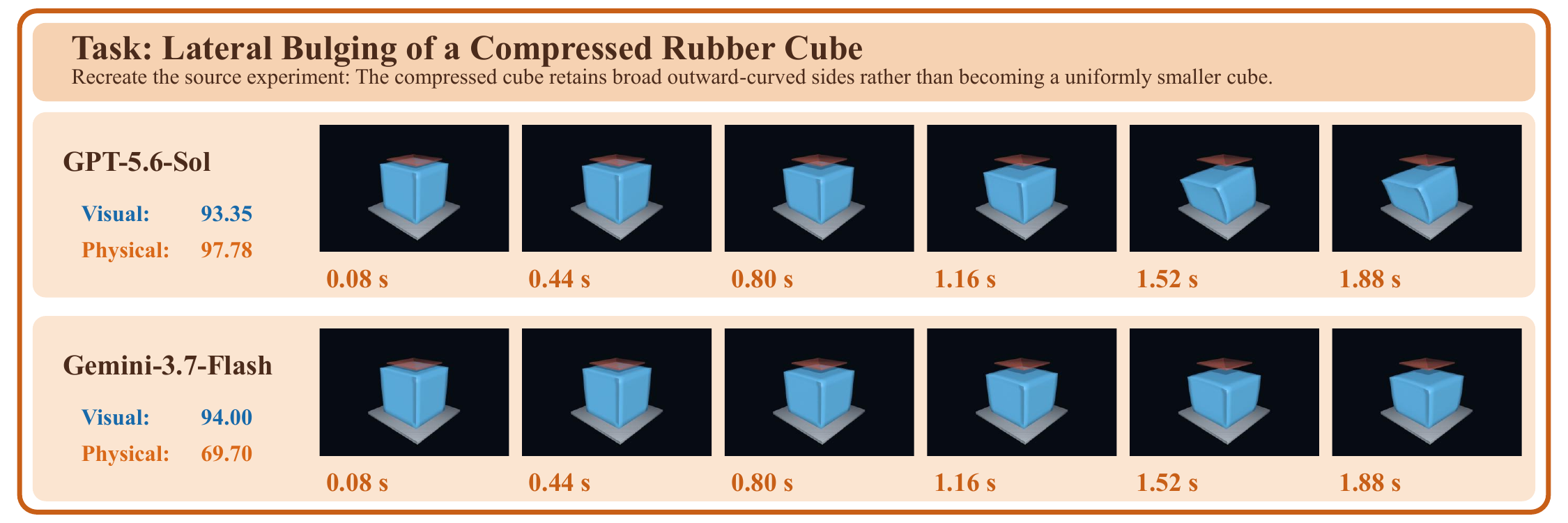}
    \caption{Case study results for the rubber cube compression task.}
    \label{fig:physical_B063}
\end{figure*}

The task shown in Figure~\ref{fig:physical_B063} simulates a rubber cube compressed vertically between a support and a loaded top face. The distinction of their physical plausibility score is primarily produced by the \texttt{quasistatic\_nodal\_force\_balance}. 
Gemini-3.7-Flash reproduces the expected macroscopic shape, but its trajectory contains a much larger node-wise force residual, which cannot be determined reliably from the rendered surface alone.

\begin{figure*}[h]
    \centering
    \includegraphics[width=0.8\textwidth]{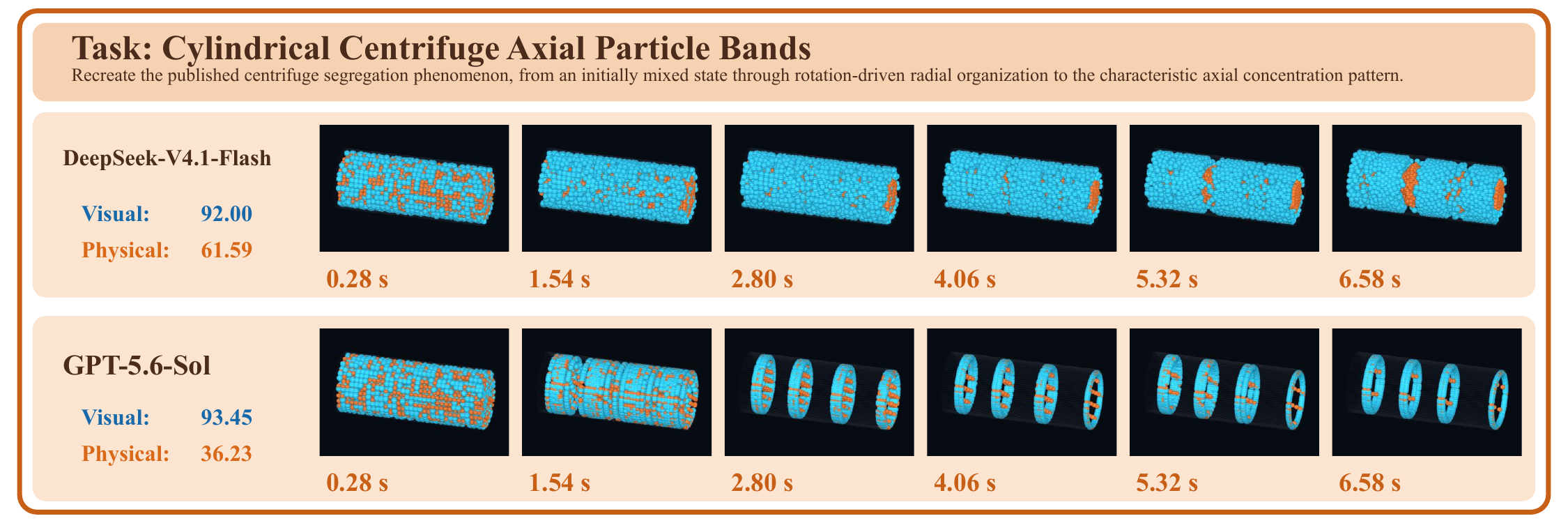}
    \caption{Case study results for the cylindrical centrifuge axial particle bands task.}
    \label{fig:physical_B165}
\end{figure*}

In figure~\ref{fig:physical_B165}, both agents simulate the characteristic axial banding pattern.
While GPT-5.6-Sol produces convincing visible bands, it fails to respect the \texttt{particle\_fluid\_force\_balance} criterion. Specifically, its exported particle-to-fluid and fluid-to-particle forces fail to cancel, revealing an inconsistent momentum exchange that is not observable in the rendered video.

\subsection{Right Physics, Wrong Appearance}
\begin{figure*}[h]
    \centering
    \includegraphics[width=0.8\textwidth]
    {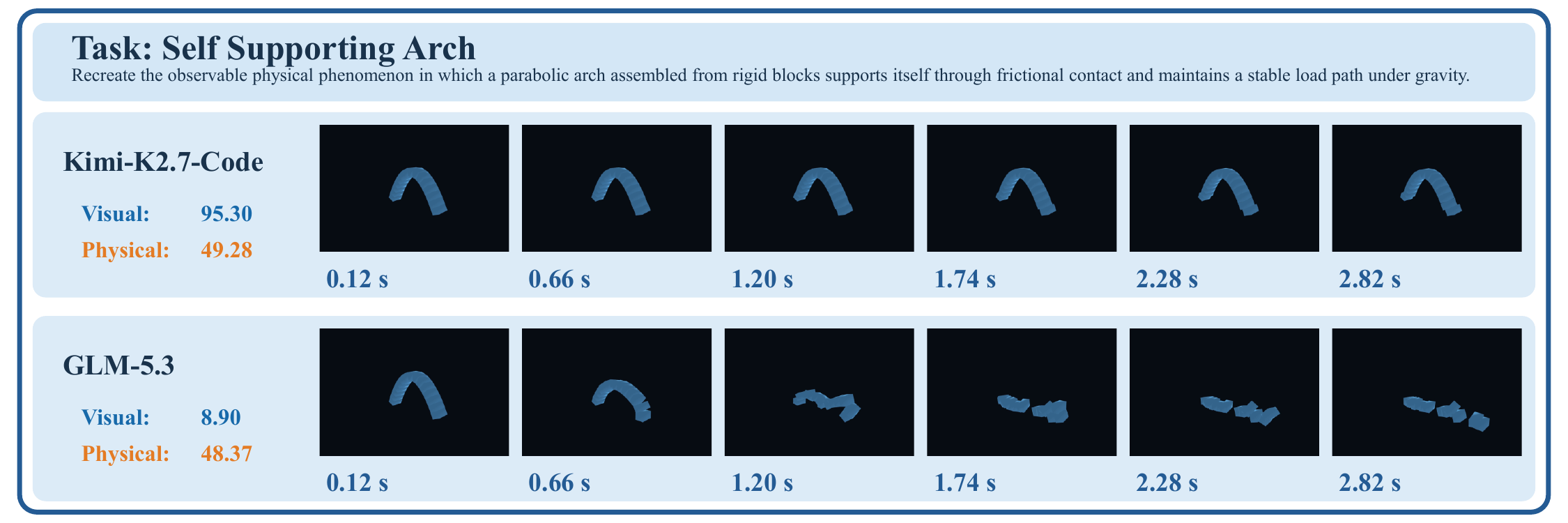}
    \caption{Case study results for the self-supporting arch task.}

    \label{fig:visual_A018}
\end{figure*}

This task simulates a parabolic arch assembled from rigid blocks that should remain standing through frictional contact. 
As shown in Figure~\ref{fig:visual_A018}, Kimi-K2.7-Code preserves the arch, whereas GLM-5.3 rapidly loses the intended structure. 
However, their physical scores are nearly identical because none of the evaluated criteria directly verifies if the complete arch remains standing. 
The visual fidelity score complements this gap by directly recognizing the stable arch as the signature outcome and strongly penalizing the collapsed scene.

\begin{figure*}[h]
    \centering
    \includegraphics[width=0.8\textwidth]
    {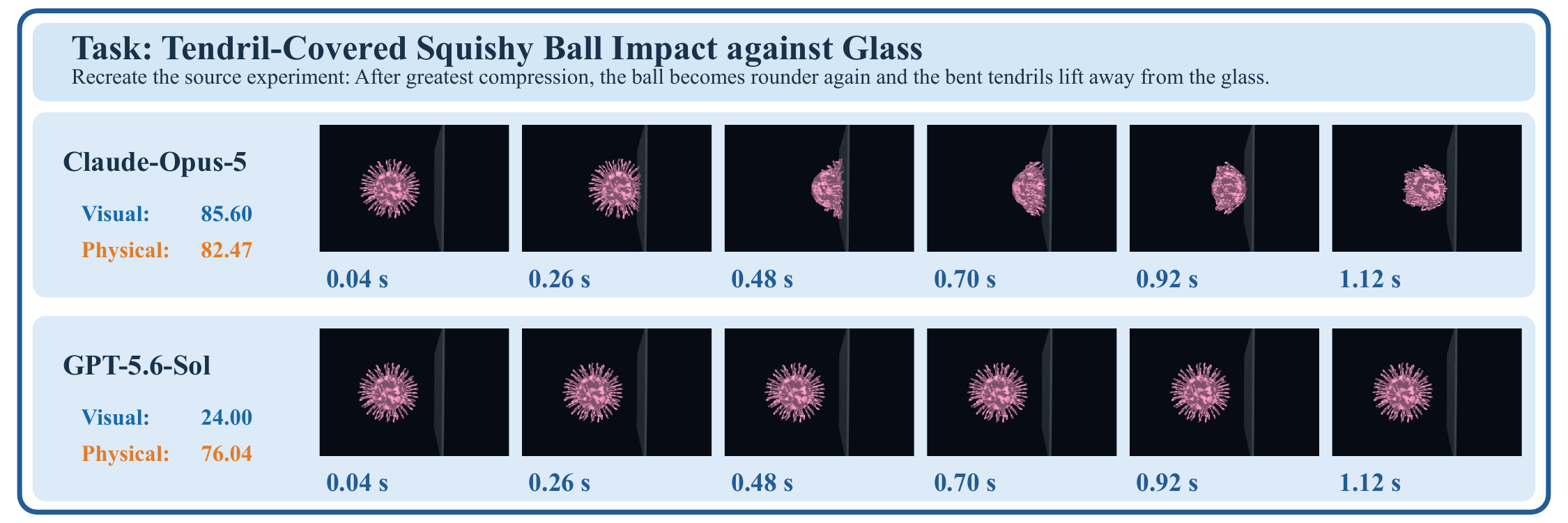}
    \caption{Case study results for the squishy ball glass impact task.}

    \label{fig:visual_A078}
\end{figure*}

In Figure~\ref{fig:visual_A078}, we provide a simulation where a tendril-covered elastic ball impacting a wall, flattening, and quickly recovering its rounded shape. 
Claude-Opus-5 reproduces the impact and recovery, whereas GPT-5.6-Sol remains almost static. Nevertheless, GPT-5.6-Sol receives a relatively high physical score because physical criteria such as \texttt{linear\_momentum\_impulse\_balance} do not directly require an impact to occur. 
A nearly static trajectory can conserve mass, avoid element inversion, and maintain a small budget residual while omitting the target phenomenon entirely.

\begin{figure*}[h]
    \centering
    \includegraphics[width=0.8\textwidth]
    {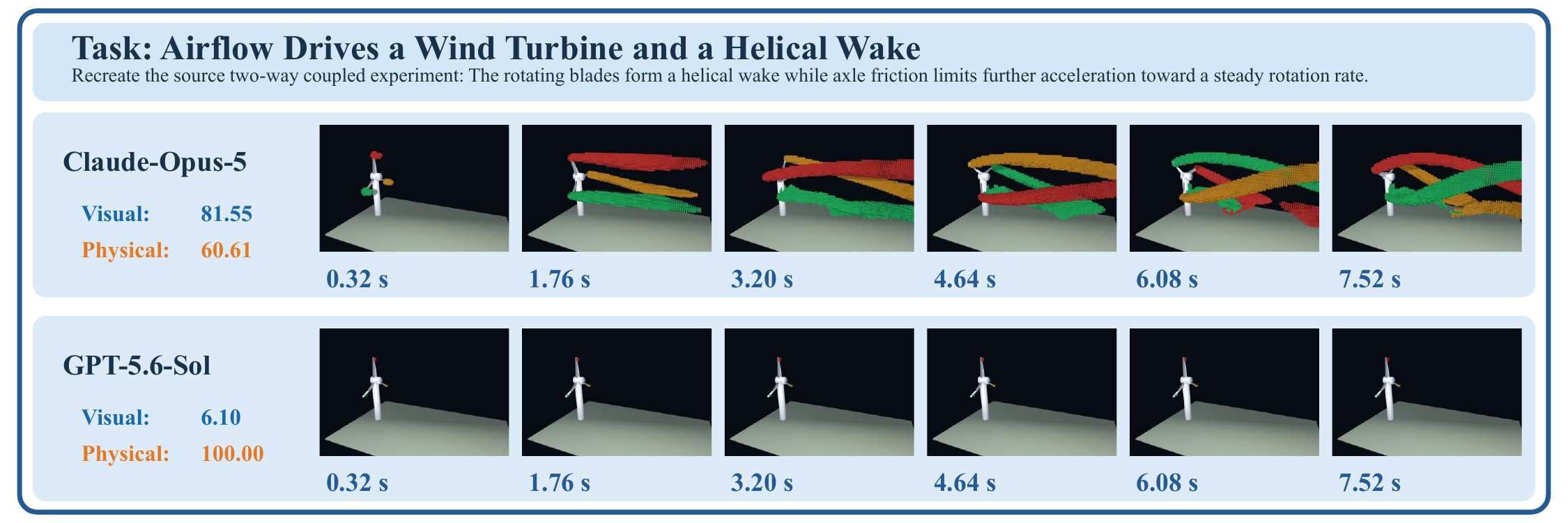}
    \caption{Case study results for the air-driven wind turbine task.}

    \label{fig:visual_A123}
\end{figure*}

As shown in Figure~\ref{fig:visual_A123}, we provide a task in which incoming airflow should accelerate a dynamic rotor and produce a visible helical smoke wake. 
GPT-5.6-Sol barely responds, yet it still receives a perfect physical score: its nearly static state drives both sides of the \texttt{angular\_momentum\_torque\_budget}, \texttt{solid\_fluid\_exchange\_force\_balance}, and \texttt{impermeable\_interface\_normal\_velocity} balances to approximately zero. 
However, visual fidelity closes this loophole, checking rotor acceleration, smoke transport, and the helical wake directly against the scene.

\begin{figure*}[h]
    \centering
    \includegraphics[width=0.8\textwidth]
    {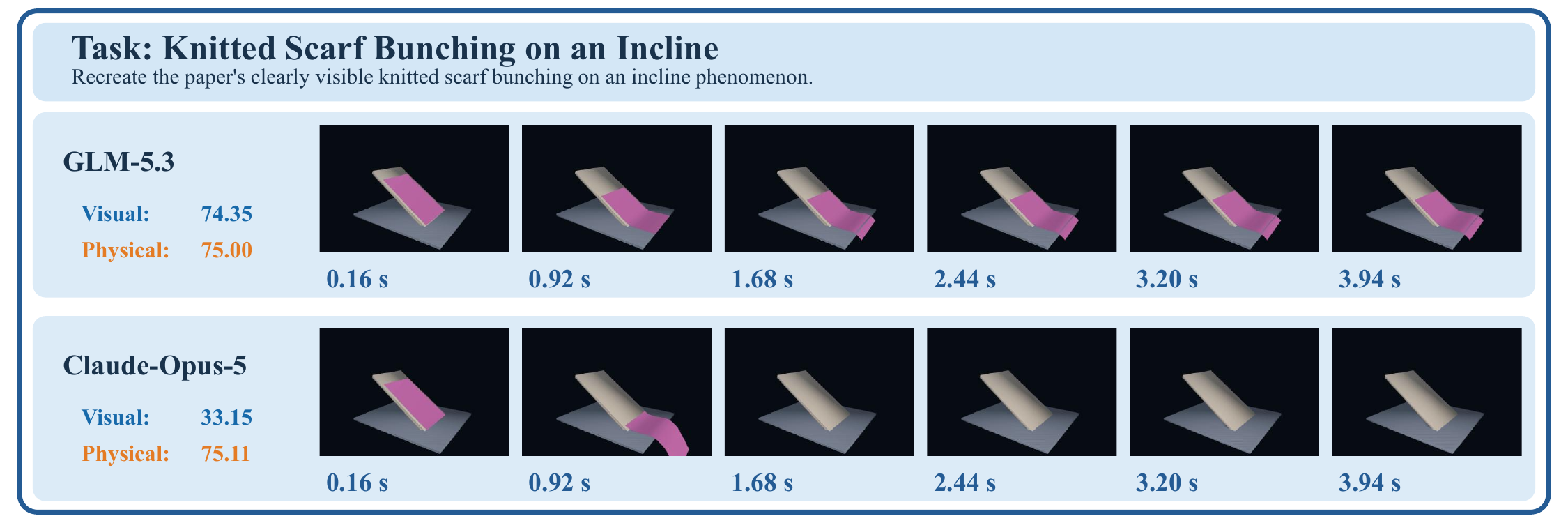}
    \caption{Case study results for the knit scarf incline bunching task.}
    \label{fig:visual_A154}
\end{figure*}

The task in Figure~\ref{fig:visual_A154} simulates a knitted scarf sliding down an incline and accumulating into thick bunched folds. 
Their physical scores are almost identical because both conserve mass, remain within the allowed surface-strain range, and avoid measured contact penetration, while both have similarly poor momentum-balance residuals. 
However, only GLM-5.3 reproduces the expected cloth pile, while in Claude-Opus-5 the scarf flies off the edge. 

\subsection{Both Metrics Agree}
As a complementary set of examples, these cases show agreement between the two metrics. Successful solvers reproduce the target phenomenon while satisfying the physical criteria, whereas unsuccessful solvers fail both visually and physically. 
Together, they illustrate that the two evaluations are consistent when solver quality isn't ambiguous.

\begin{figure*}[h]
    \centering
    \includegraphics[width=0.8\textwidth]
    {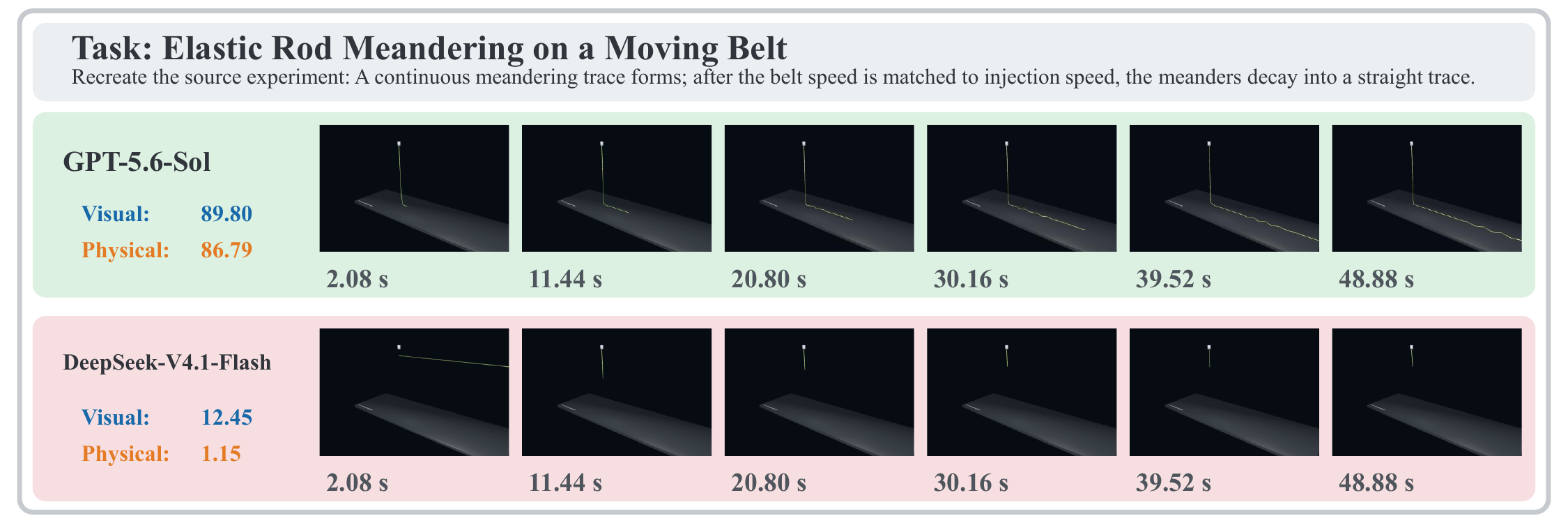}

    \centering
    \includegraphics[width=0.8\textwidth]
    {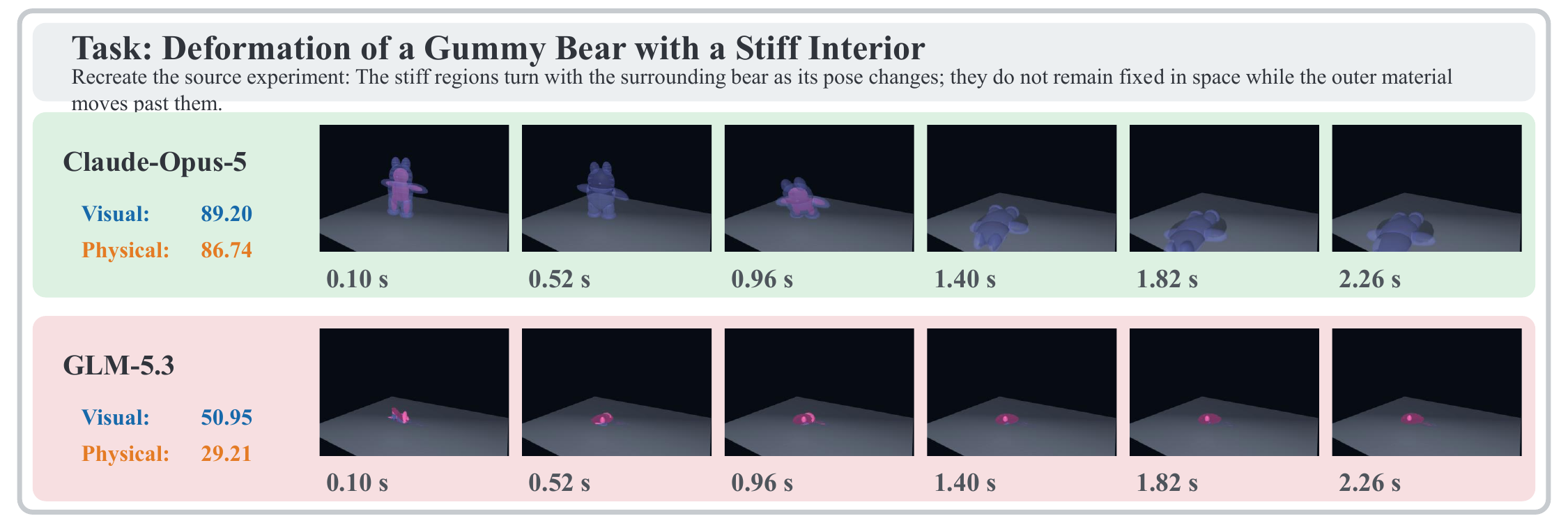}

    \centering
    \includegraphics[width=0.8\textwidth]
    {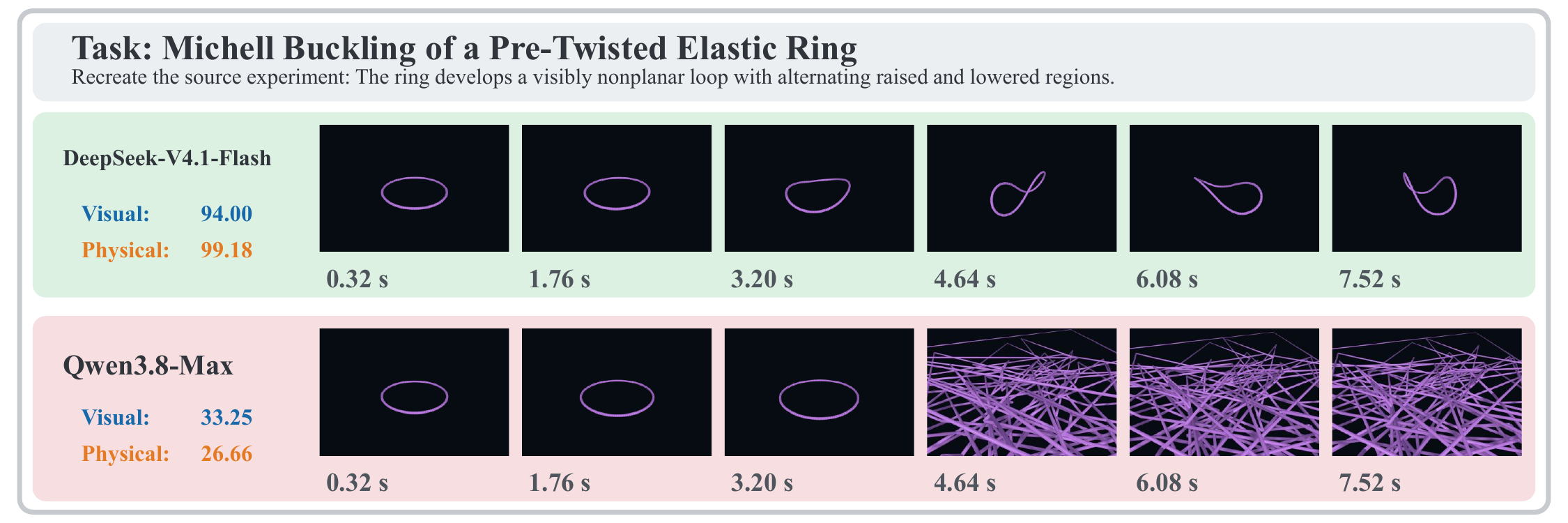}

    \centering
    \includegraphics[width=0.8\textwidth]
    {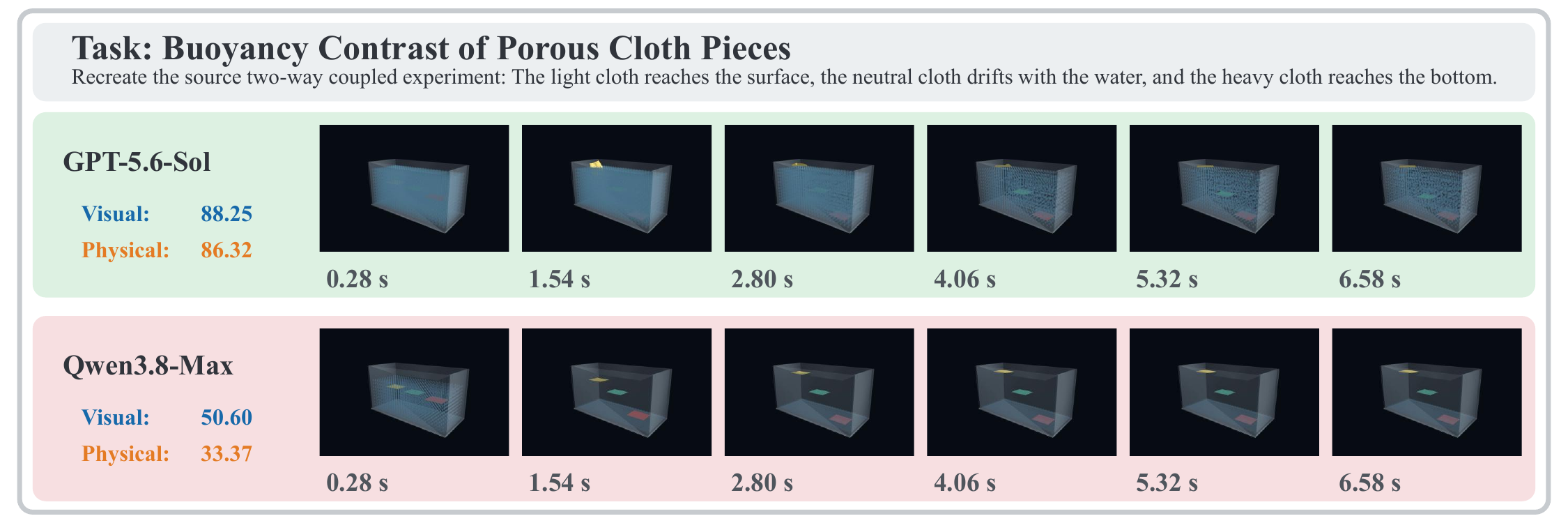}
    \caption{Case study results for unambiguous solvers.}
   
\end{figure*}

\section{More Results}
\label{app:more_results}
\subsection{Complete Main Results}
\begin{figure*}[p]
    \centering

    % ================= Row 1 =================
    \begin{subfigure}[t]{0.48\textwidth}
        \centering
        \includegraphics[width=\linewidth]
        {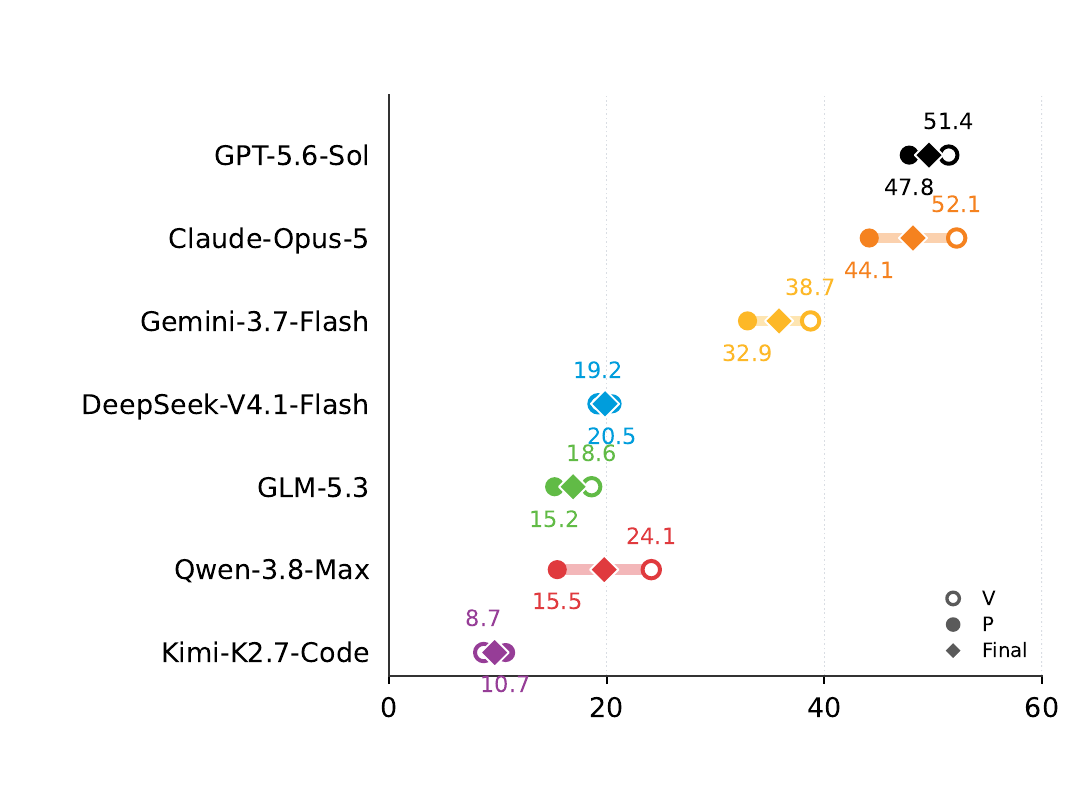}
        \caption{Fluid dynamics}
    \end{subfigure}
    \hfill
    \begin{subfigure}[t]{0.48\textwidth}
        \centering
        \includegraphics[width=\linewidth]
        {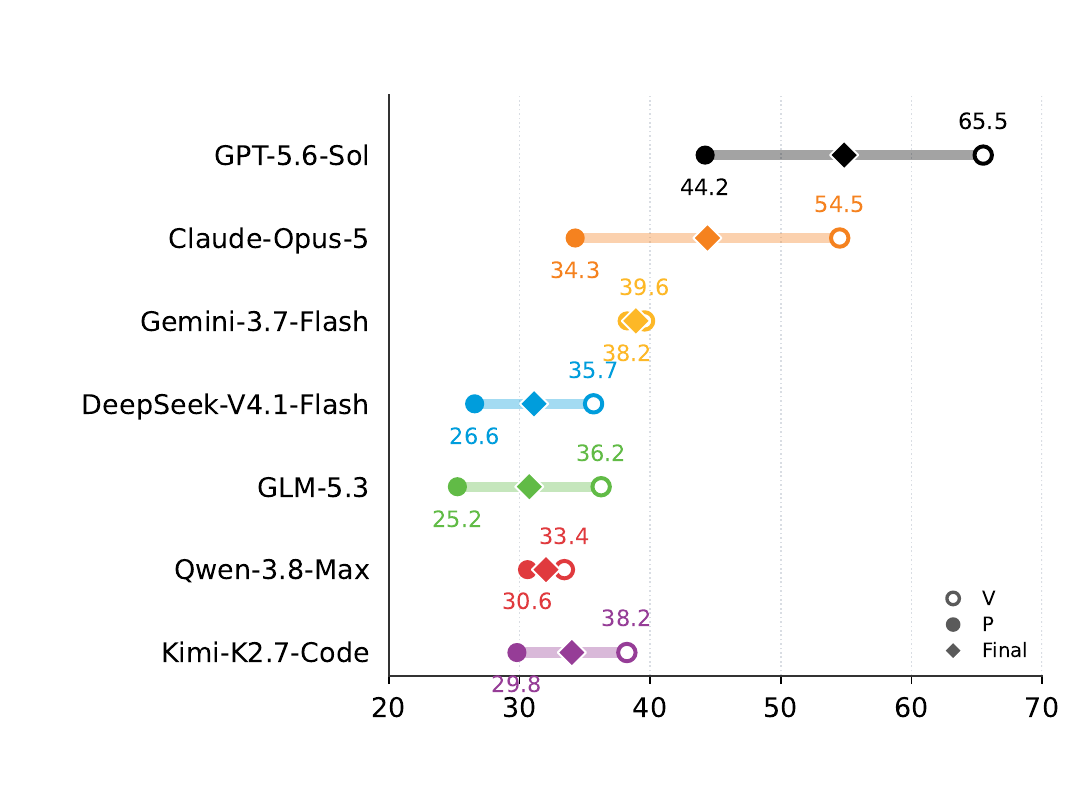}
        \caption{Rigid body dynamics}
    \end{subfigure}

    \vspace{0.8em}

    % ================= Row 2 =================
    \begin{subfigure}[t]{0.48\textwidth}
        \centering
        \includegraphics[width=\linewidth]
        {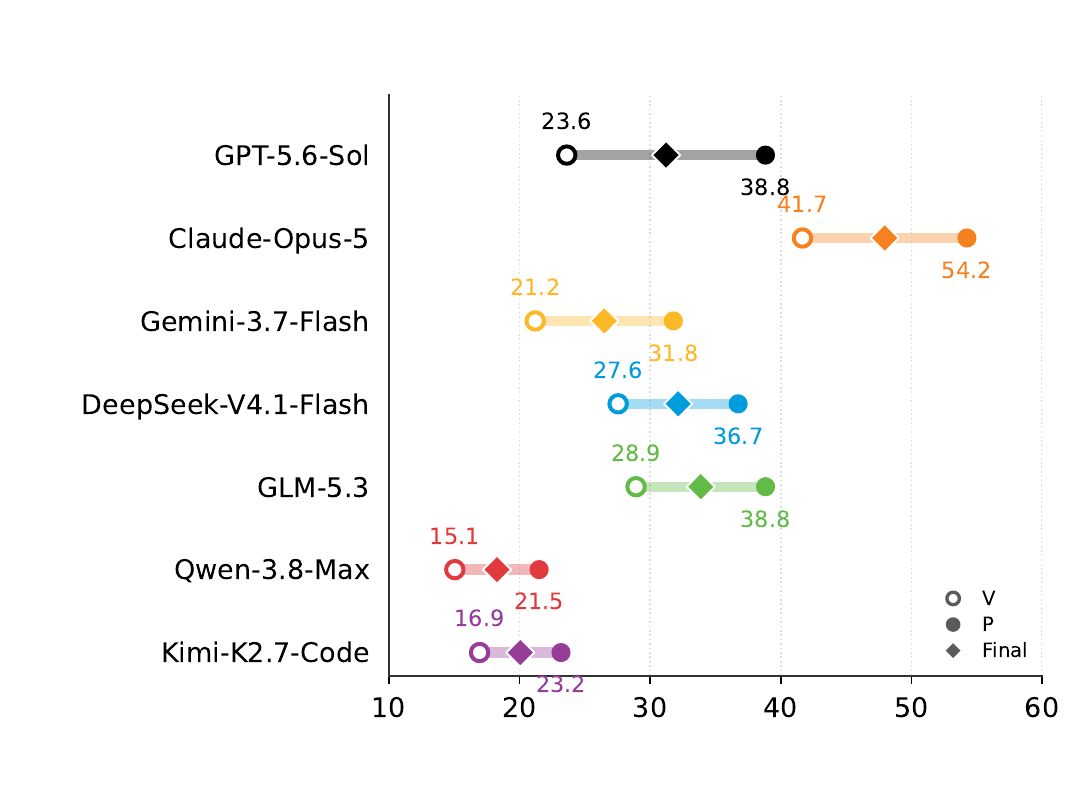}
        \caption{Plastic and complex materials}
    \end{subfigure}
    \hfill
    \begin{subfigure}[t]{0.48\textwidth}
        \centering
        \includegraphics[width=\linewidth]
        {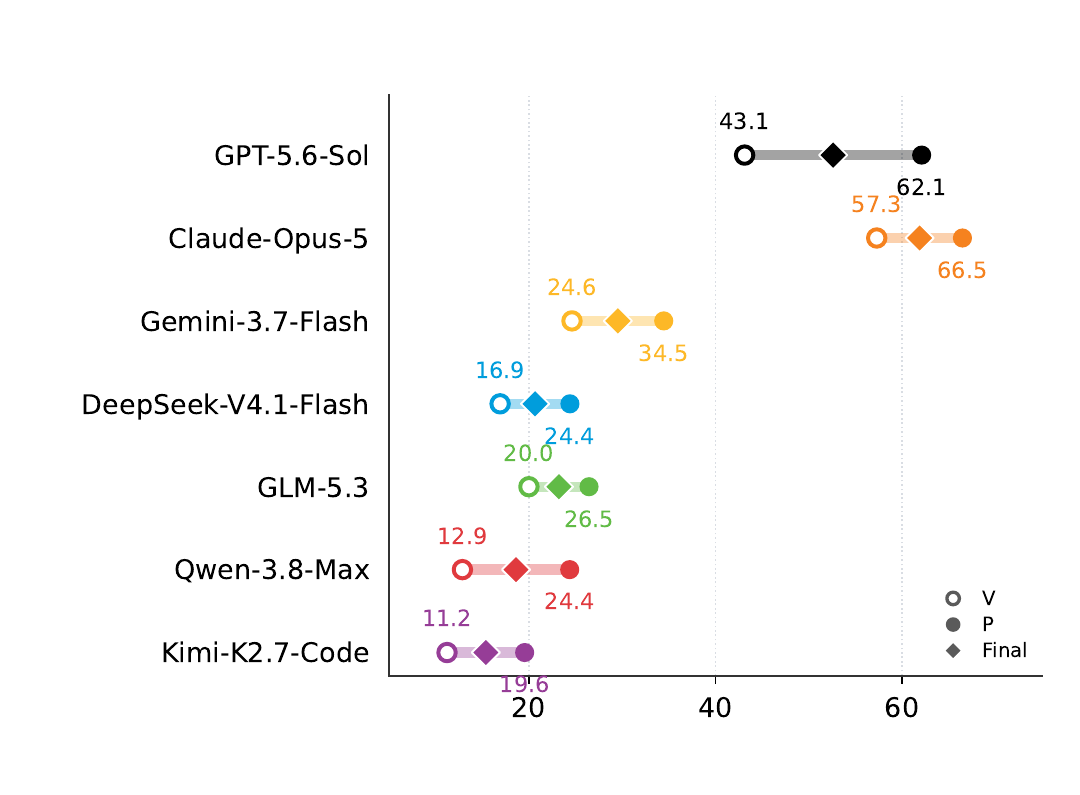}
        \caption{Elastic solids}
    \end{subfigure}

    \vspace{0.8em}

    % ================= Row 3 =================
    \begin{subfigure}[t]{0.48\textwidth}
        \centering
        \includegraphics[width=\linewidth]
        {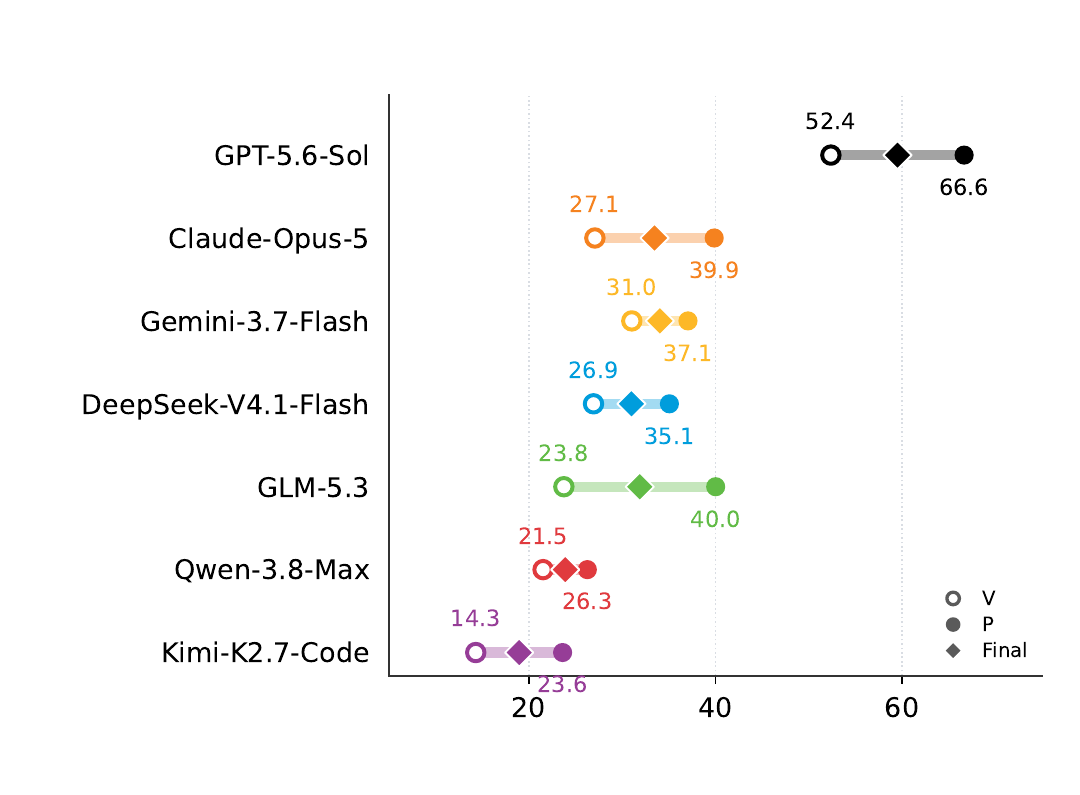}
        \caption{Rods and strands}
    \end{subfigure}
    \hfill
    \begin{subfigure}[t]{0.48\textwidth}
        \centering
        \includegraphics[width=\linewidth]
        {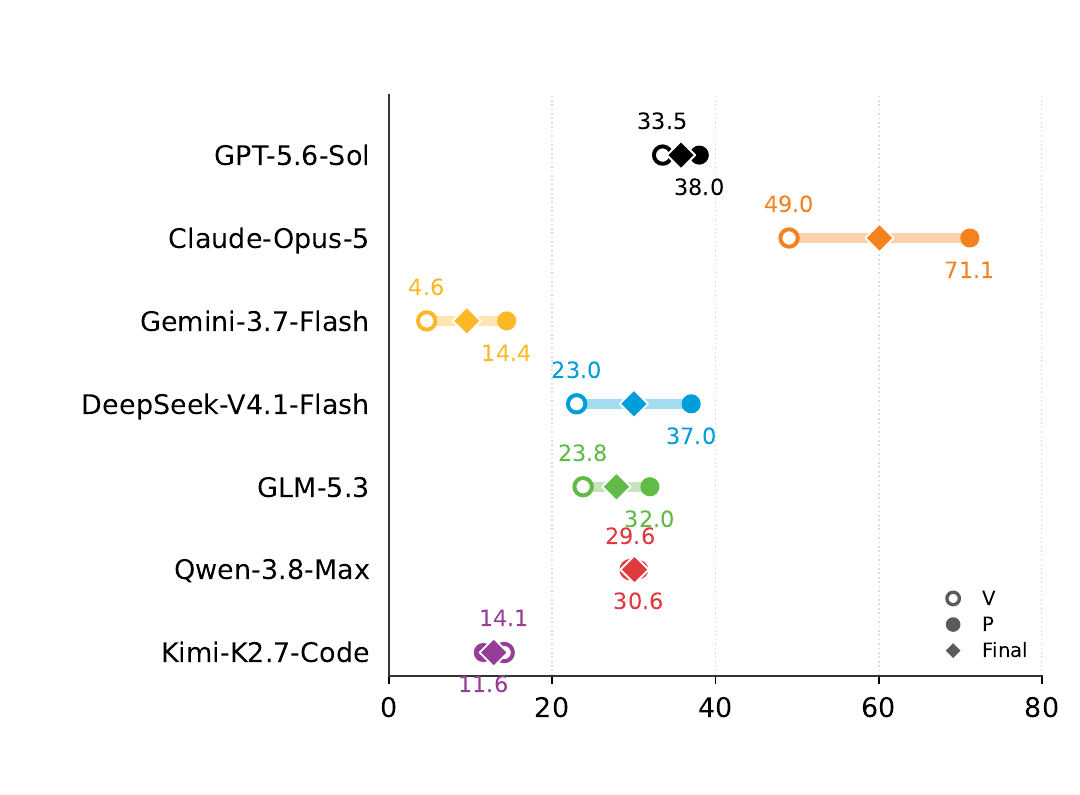}
        \caption{Cloth and shells}
    \end{subfigure}

    \vspace{0.8em}

    % ================= Row 4 =================
    \begin{subfigure}[t]{0.48\textwidth}
        \centering
        \includegraphics[width=\linewidth]
        {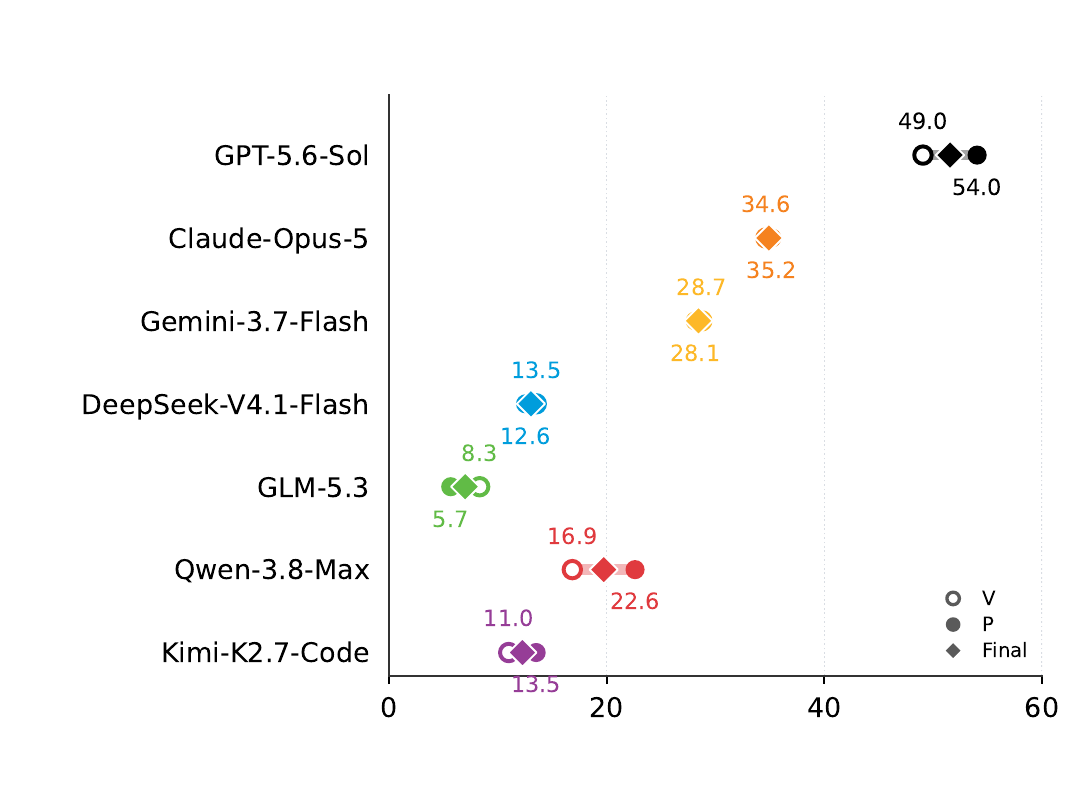}
        \caption{Multiphysics coupling}
    \end{subfigure}
    \hfill
    \begin{subfigure}[t]{0.48\textwidth}
        \centering
        \includegraphics[width=\linewidth]
        {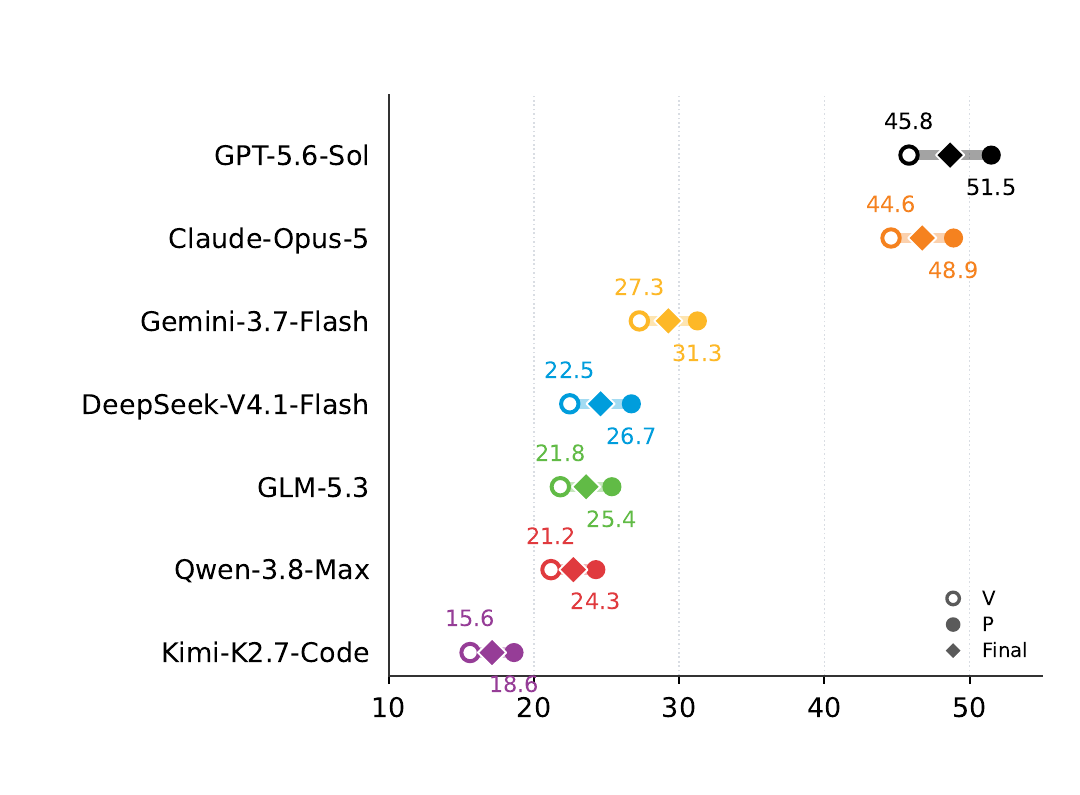}
        \caption{Overall}
    \end{subfigure}

    \caption{Visual Fidelity and Physical Plausibility scores across different physical domains.}
    \label{fig:vp_domains}
\end{figure*}

Figure~\ref{fig:vp_domains} shows broadly consistent performance trends between visual and physics scores, together with systematic differences across domains.
In rigid-body dynamics, every model scores higher visually than physically. 
For example, GPT-5.6-Sol obtains 65.5\% versus 44.2\%.
This gap suggests that visually plausible motion can coexist with violations detected by the physical checks, demonstrating the necessity of physical evaluation.
The pattern reverses for plastic and complex materials, where physics scores consistently exceed visual scores, suggesting that, in certain domains, even minor violations of physical constraints by a solver can lead to poor visual results.
The two criteria therefore provide complementary evidence:
visual evaluation assesses observable fidelity, while physics evaluation checks constraints that plausible-looking outputs may nevertheless violate.

\subsection{Visual-Judge Stability}
\begin{figure*}[t]
    \centering

    % ================= First row =================
    \begin{minipage}[t]{0.48\textwidth}
        \centering
        \includegraphics[width=\linewidth]
        {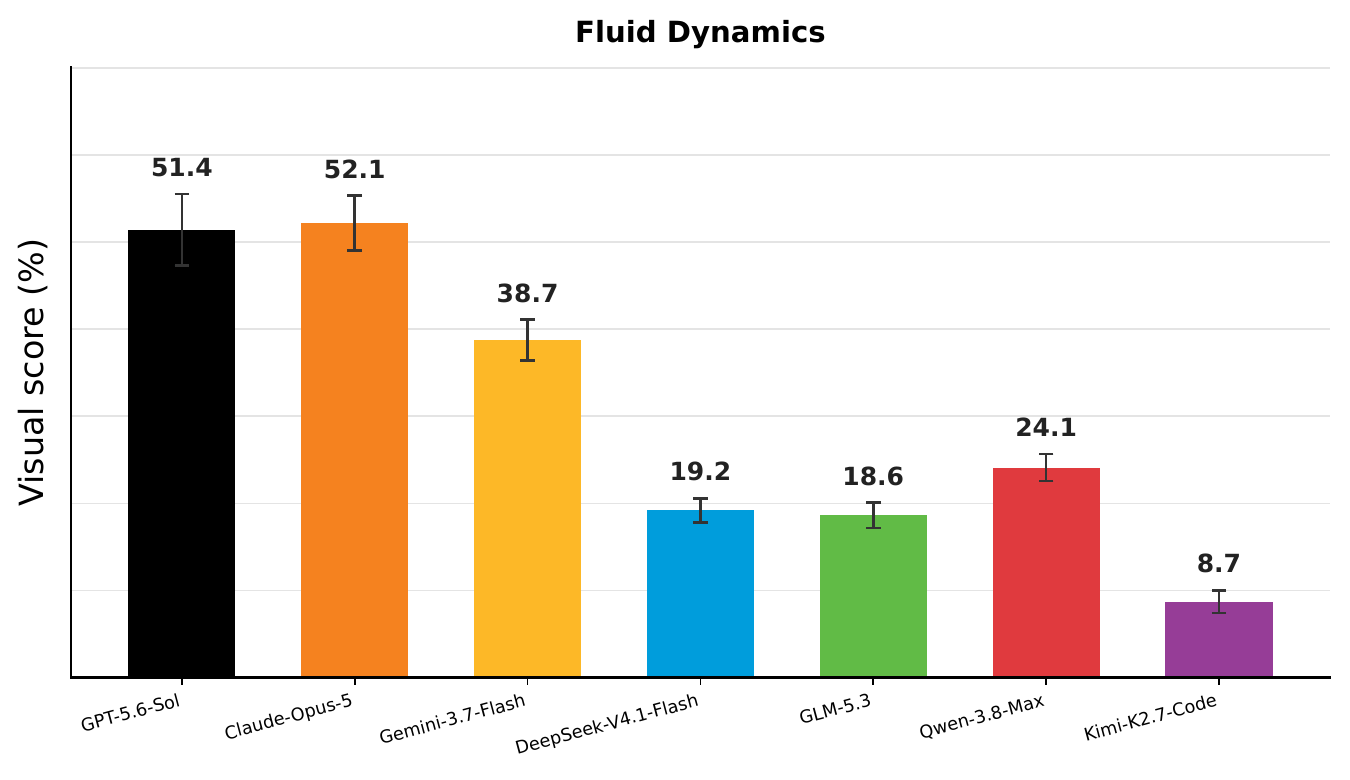}

        \vspace{1mm}
        {\small Fluid}
    \end{minipage}
    \hfill
    \begin{minipage}[t]{0.48\textwidth}
        \centering
        \includegraphics[width=\linewidth]
        {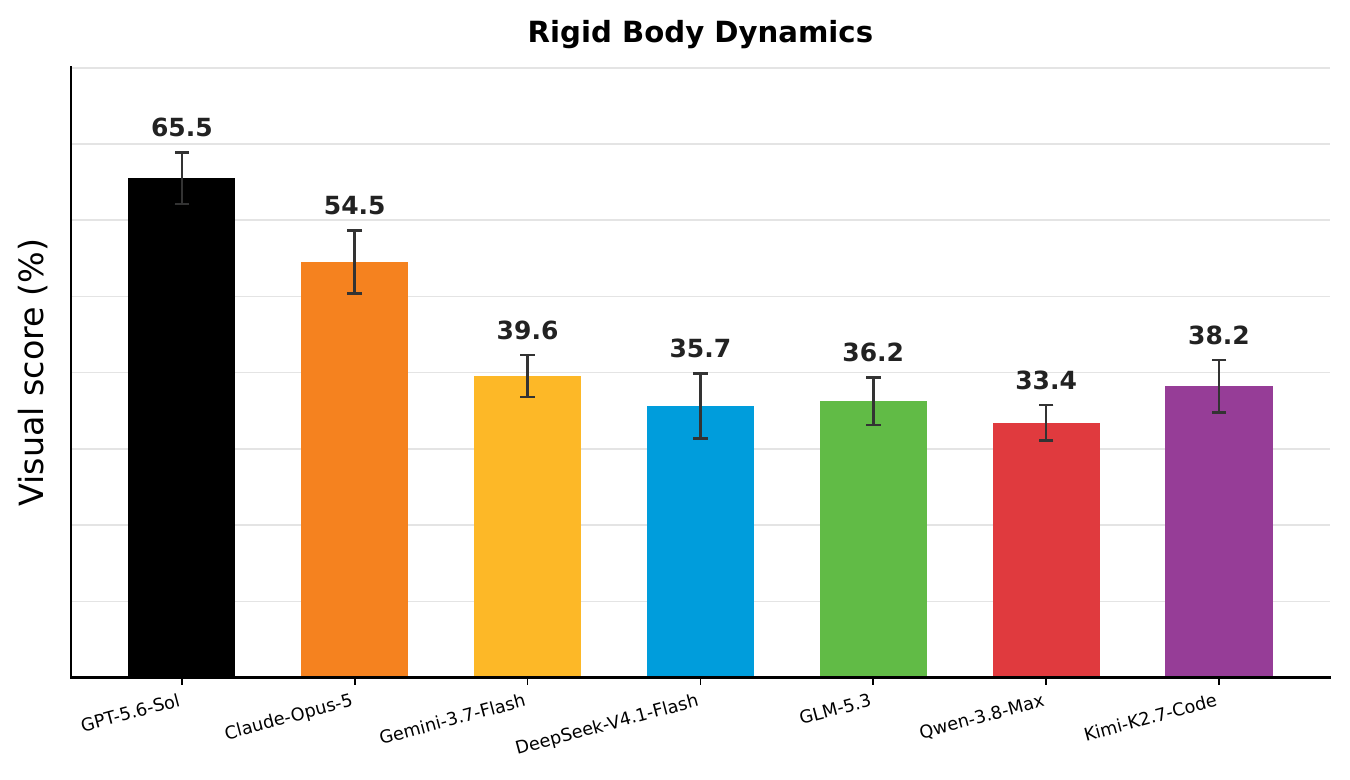}

        \vspace{1mm}
        {\small Rigid Body}
    \end{minipage}

    \vspace{3mm}

    % ================= Second row =================
    \begin{minipage}[t]{0.48\textwidth}
        \centering
        \includegraphics[width=\linewidth]
        {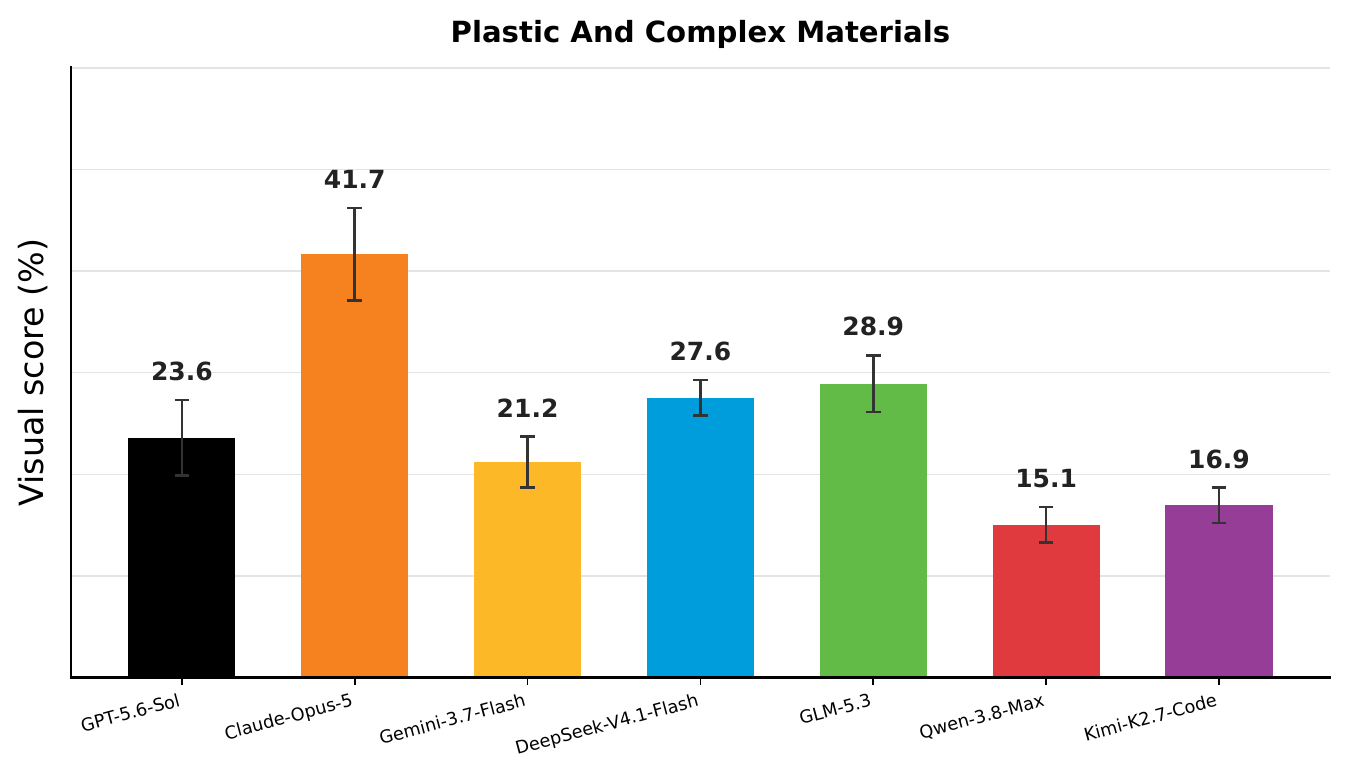}

        \vspace{1mm}
        {\small Plastic \& Complex Materials}
    \end{minipage}
    \hfill
    \begin{minipage}[t]{0.48\textwidth}
        \centering
        \includegraphics[width=\linewidth]
        {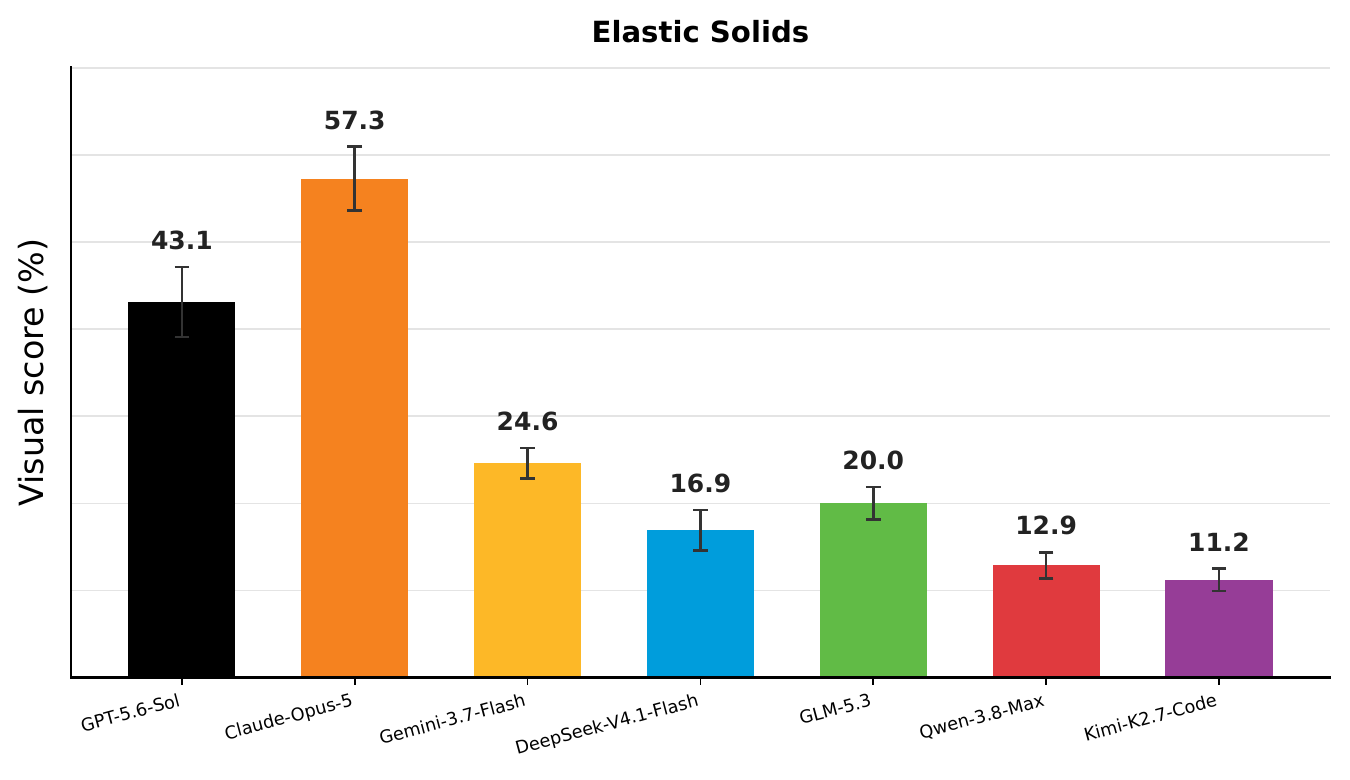}

        \vspace{1mm}
        {\small Elastic Solids}
    \end{minipage}

    \vspace{3mm}

    % ================= Third row =================
    \begin{minipage}[t]{0.48\textwidth}
        \centering
        \includegraphics[width=\linewidth]
        {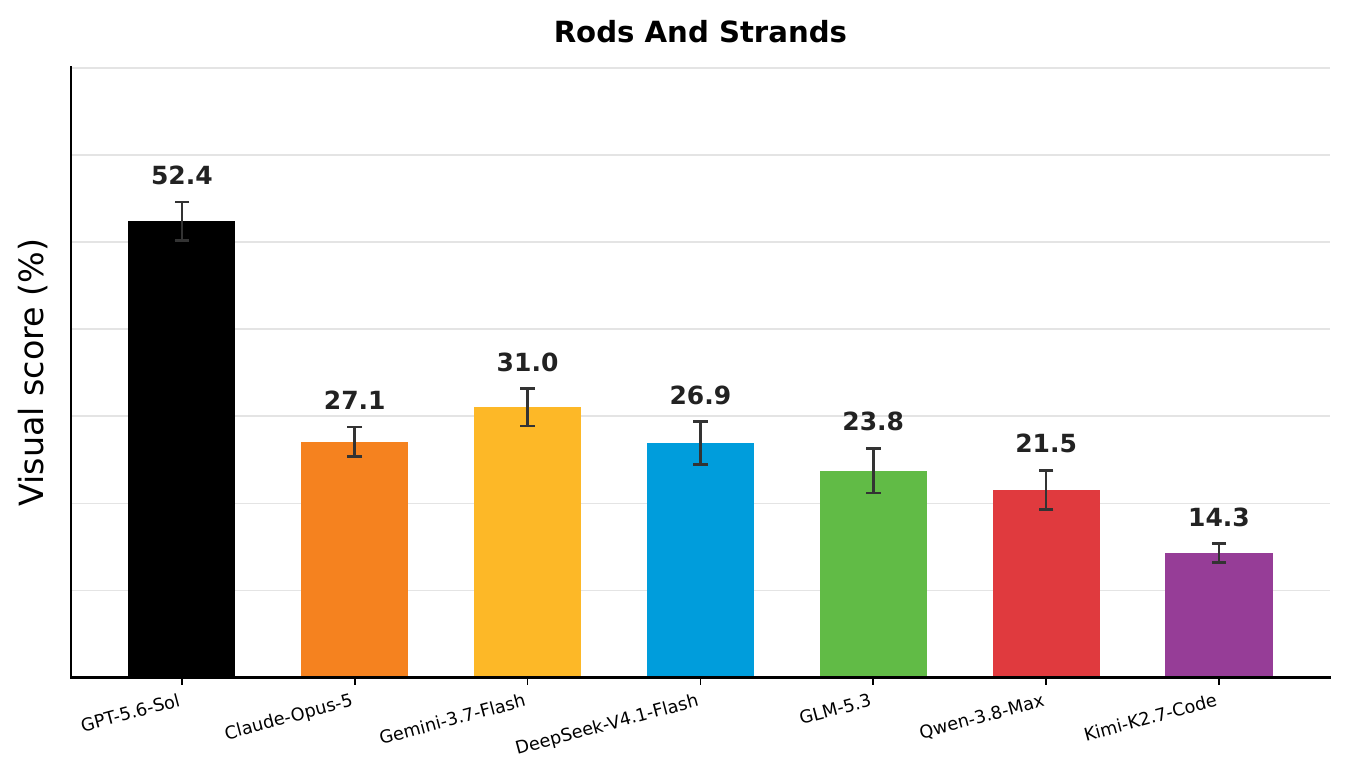}

        \vspace{1mm}
        {\small Rods \& Strands}
    \end{minipage}
    \hfill
    \begin{minipage}[t]{0.48\textwidth}
        \centering
        \includegraphics[width=\linewidth]
        {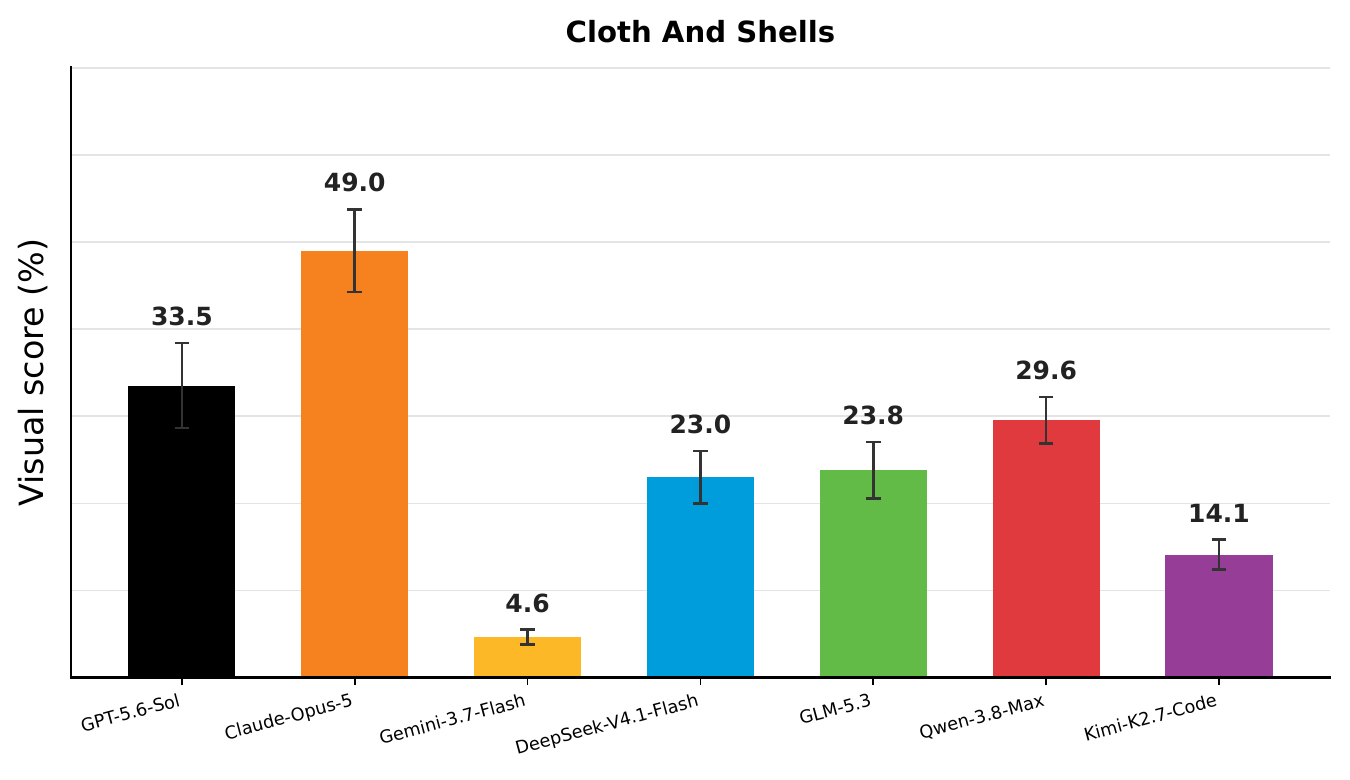}

        \vspace{1mm}
        {\small Cloth \& Shells}
    \end{minipage}

    \vspace{3mm}

    % ================= Fourth row =================
    \begin{minipage}[t]{0.48\textwidth}
        \centering
        \includegraphics[width=\linewidth]
        {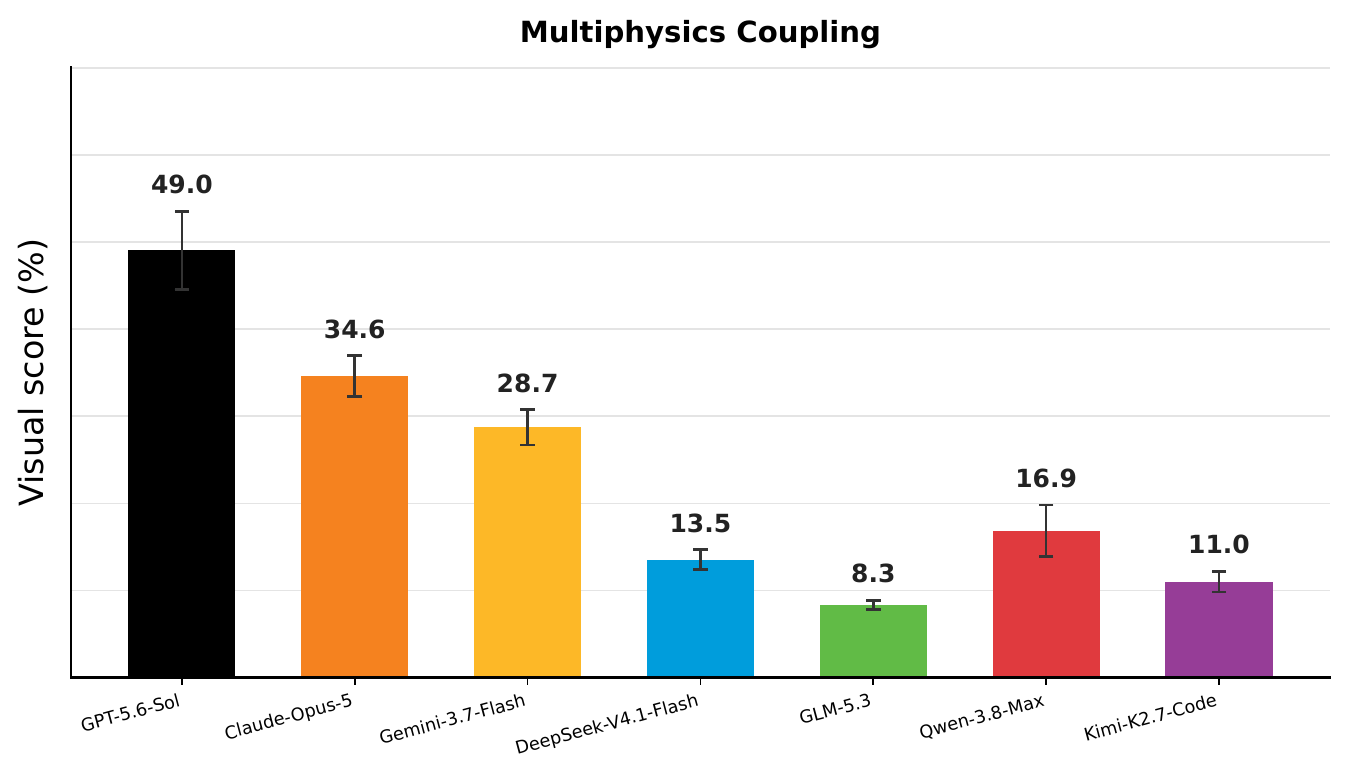}

        \vspace{1mm}
        {\small Multiphysics Coupling}
    \end{minipage}
    \hfill
    \begin{minipage}[t]{0.48\textwidth}
        \centering
        \includegraphics[width=\linewidth]
        {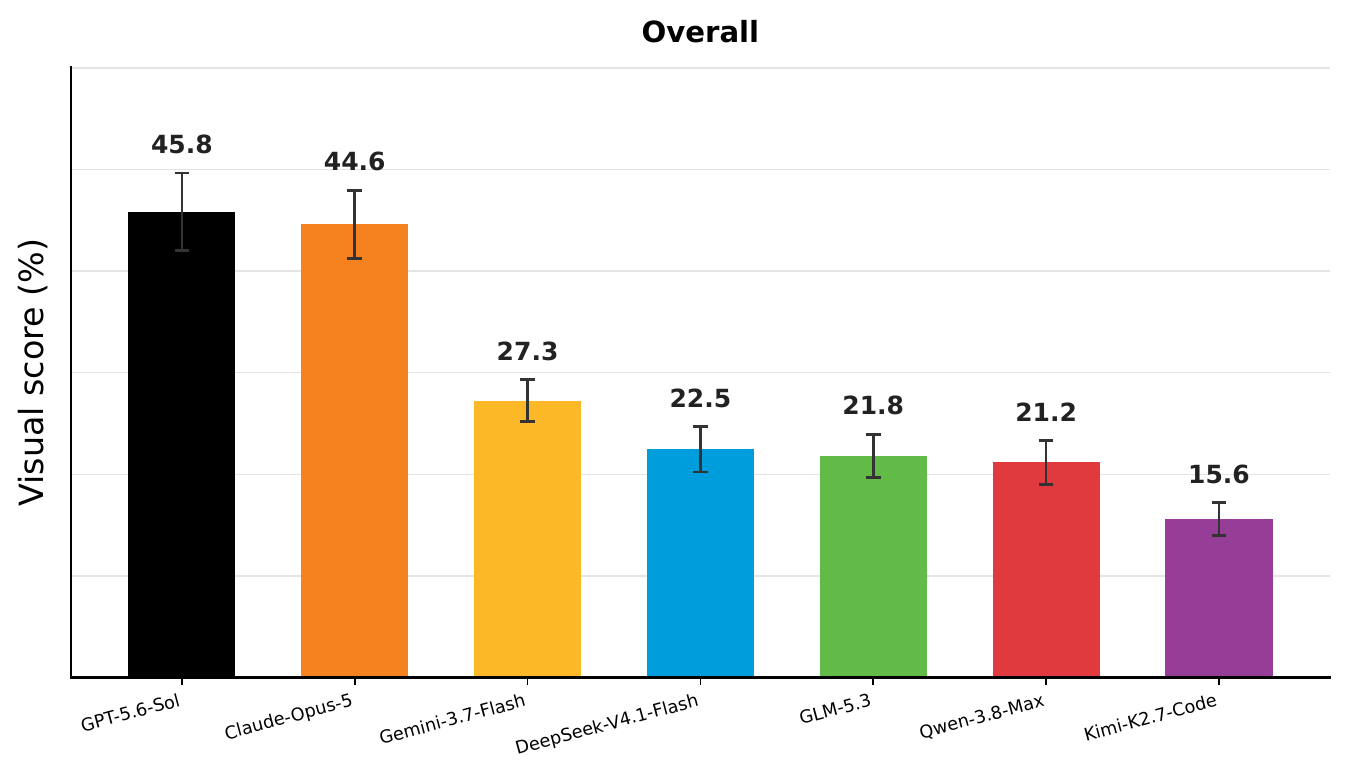}

        \vspace{1mm}
        {\small Overall}
    \end{minipage}

    \caption{Performance across different physics domains.}
    \label{fig:error_bar}
\end{figure*}

To verify the stability of the VLM judge, we repeat the visual-fidelity evaluation of each task five times.
Figure~\ref{fig:error_bar} reports the mean scores per domain, with error bars indicating the mean standard deviation within the task in the five evaluations.
The error bars are generally small across agents and physical domains, showing that repeated evaluations produce consistent scores.
Slightly larger variations appear in some cases, but the overall performance patterns remain stable.
These results suggest that our VLM judge provides consistent visual-fidelity evaluations reliably.

\subsection{Human Annotations}
Our Visual Fidelity evaluation consists of four task-specific criteria for each case, which are assessed by a VLM using a shared set of key frames.
To validate this evaluation protocol against human judgment, we asked expert annotators to assess solver quality using the same key frames and evaluation criteria.
Among the 661 cases with successfully rendered videos, we sampled 231 cases for human evaluation, including 40 overlapping cases rated by multiple annotators to measure Human-Human consistency.
As shown in Table~\ref{tab:human_vlm_correlation}, the VLM scores show strong agreement with human judgments in both absolute values (Pearson $r=0.730$) and rank ordering (Spearman $\rho=0.713$).
For comparison, Human-Human correlations on the shared cases are lower, with Pearson $r=0.624$ and Spearman $\rho=0.607$.
These results indicate that the VLM-based evaluation aligns well with expert judgments and achieves correlations, in this sample higher, than those observed between individual human annotators.
\begin{table}[h]
\centering
\caption{Correlation of VLM and human judgments.}
\label{tab:human_vlm_correlation}
\begin{tabular}{lcc}
\toprule
Comparison & Pearson $r$ & Spearman $\rho$ \\
\midrule
Human--VLM   & 0.730 & 0.713 \\
Human--Human  & 0.624 & 0.607 \\
\bottomrule
\end{tabular}
\end{table}

\subsection{Solver Style}
To grasp a view of each agent's output solver style, we further provide the size and runtime efficiency of the generated solvers in Table~\ref{tab:solver_statistics}.
Being the best performance model, GPT-5.6-Sol also produces the most compact implementations, averaging 234 lines and 1,028 words, while achieving the shortest execution time of 22.78 seconds.
Though showing comparable performance, Claude-Opus-5 attains a comparable benchmark score, but its solvers are the longest on average and have the highest mean execution time (100.79 seconds).
GLM-5.3 has the lowest mean memory consumption (235.81 MiB), but its solvers are longer than those of GPT-5.6-Sol in context.
Overall, solver size and resource usage vary substantially across agents, even when their performance scores are similar.

\clearpage
\definecolor{bestColor}{HTML}{FFF3CC}

\begin{table*}[h]
\centering
\renewcommand{\arraystretch}{1.15}
\caption{Statistics of generated solvers and their runtime resource usage. 
We report the mean and maximum solver size in lines and words, together with execution time and memory consumption. 
The minimum value in each column is highlighted \protect\colorbox{bestColor}{\strut yellow}.}
\label{tab:solver_statistics}
\resizebox{0.95\textwidth}{!}{%
\begin{tabular}{lcccccc}
    \toprule
    \multirow{2}{*}{Model}
    & \multicolumn{2}{c}{Lines}
    & \multicolumn{2}{c}{Words}
    & \multicolumn{2}{c}{Runtime} \\
    \cmidrule(lr){2-3}
    \cmidrule(lr){4-5}
    \cmidrule(lr){6-7}
    & Mean & Max
    & Mean & Max
    & Time (s) & Memory (MiB) \\
    \midrule

    GPT-5.6 Sol~\citep{openaiGpt56}
    & \cellcolor{bestColor}234
    & \cellcolor{bestColor}606
    & \cellcolor{bestColor}1028
    & \cellcolor{bestColor}2765
    & \cellcolor{bestColor}22.78
    & 388.21 \\

    Claude Opus 5~\citep{anthropicOpus5}
    & 523 & 878 & 2382 & 4183 & 100.79 & 347.28 \\

    Gemini 3.7 Flash~\citep{googleGemini37Flash}
    & 345 & 639 & 1495 & 3130 & 61.29 & 414.26 \\

    DeepSeek-V4.1-Flash~\citep{deepseekV41Flash}
    & 512 & 1060 & 2293 & 4951 & 77.04 & 355.77 \\

    GLM-5.3~\citep{zaiGlm53}
    & 440 & 1014 & 2021 & 5148 & 59.42
    & \cellcolor{bestColor}235.81 \\

    Qwen-3.8-Max~\citep{alibabaQwen38Max}
    & 371 & 806 & 1705 & 3522 & 68.09 & 258.59 \\

    Kimi-K2.7-Code~\citep{moonshotKimiK27}
    & 372 & 707 & 1496 & 2868 & 60.94 & 295.42 \\

    \bottomrule
\end{tabular}%
}
\end{table*}

\subsection{Task-Level Results}
\begin{figure}
    \centering
    \includegraphics[width=0.75\linewidth]{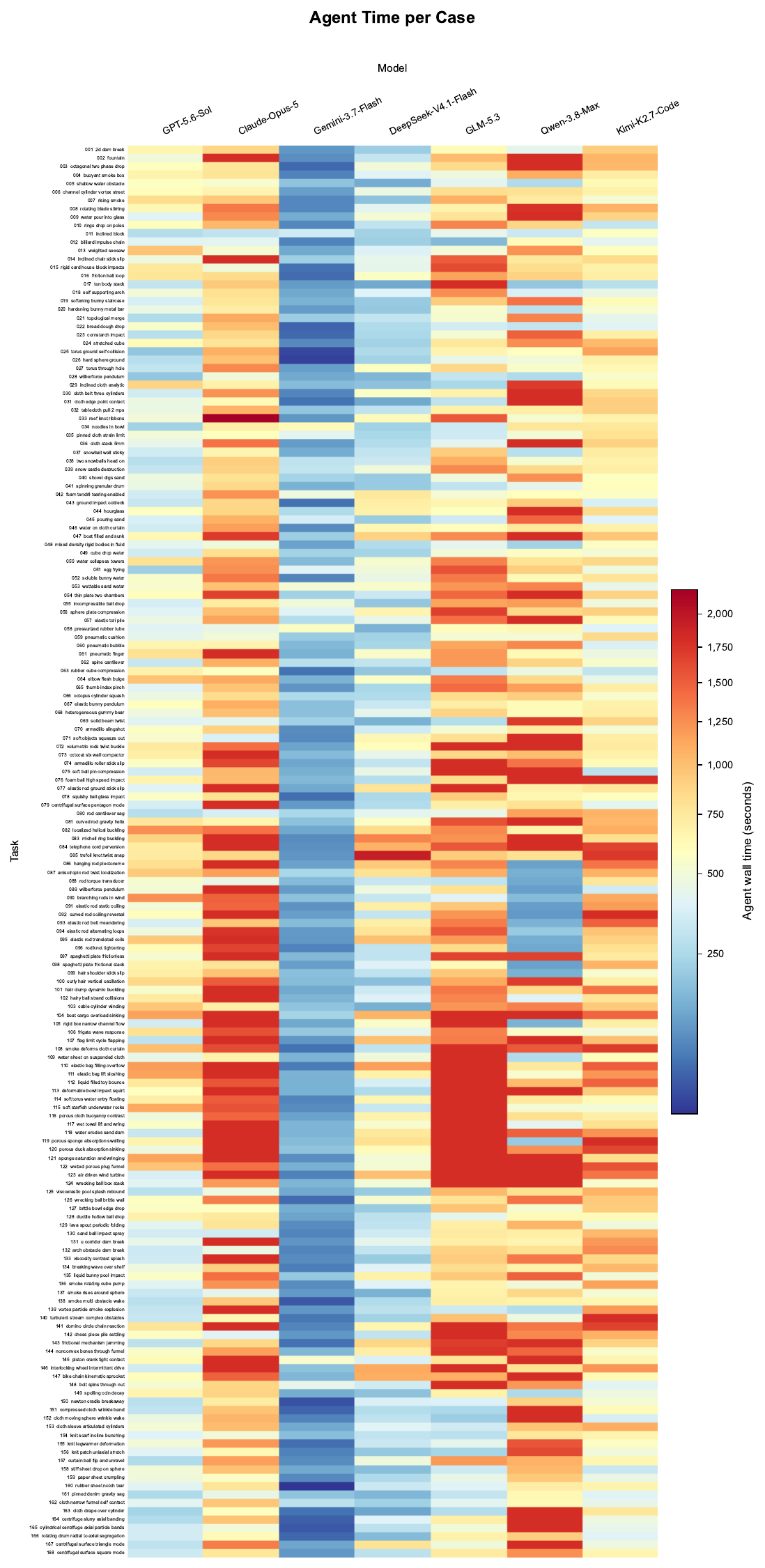}
    \caption{Per-case agent wall time across 168 benchmark tasks and seven models. Each cell represents the generation time consumed by the agent in a single run. Blue indicates shorter durations, while red indicates longer durations.}
    \label{fig:heat_time}
\end{figure}

\begin{figure}
    \centering
    \includegraphics[width=0.75\linewidth]{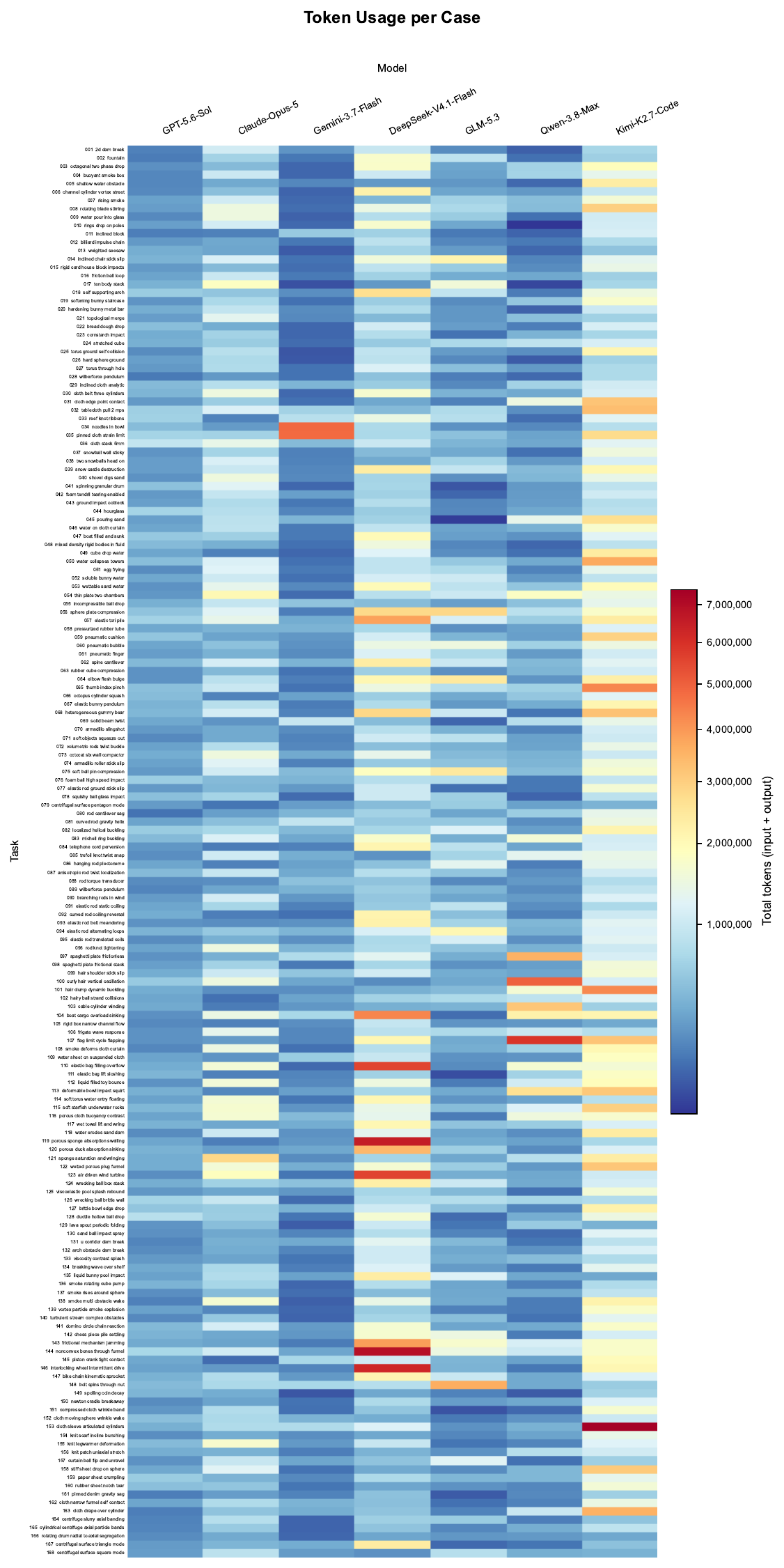}
    \caption{Per-case token usage across 168 benchmark tasks and seven models. Each cell represents the total input and output tokens consumed in a single run. Blue indicates lower usage, while red indicates higher usage.}
    \label{fig:heat_token}
\end{figure}

Figures~\ref{fig:heat_time} and~\ref{fig:heat_token} report per-task agent wall time and token usage, respectively.
The heatmaps reveal substantial variation across both agents and tasks: some agents consistently consume more time or tokens, while certain tasks require considerably greater resources across multiple agents.
These task-level results suggest that the computational difficulty of solver generation is highly uneven and depends on both the physical phenomena and the agent's problem-solving ability.

\end{document}